\documentclass{article}

\usepackage[numbers]{natbib}
\usepackage[preprint]{neurips_2022}

\usepackage[dvipsnames]{xcolor}         
\definecolor{linkColor}{rgb}{0.2,0.4,0.6}
\usepackage[utf8]{inputenc} 
\usepackage[T1]{fontenc}    
\usepackage[colorlinks=true,linkcolor=linkColor,citecolor=linkColor,filecolor=linkColor,urlcolor=linkColor]{hyperref}       
\usepackage{url}            
\usepackage{booktabs}       
\usepackage{amsfonts}       
\usepackage{nicefrac}       
\usepackage{graphicx}
\usepackage{makecell}
\usepackage{bbm}
\usepackage{subcaption}
\usepackage{soul}
\usepackage{fontawesome5}
\usepackage{pifont}
\usepackage{multirow}
\usepackage{mathtools}
\usepackage{enumitem}
\usepackage{adjustbox}
\usepackage{setspace}
\usepackage{ulem}
\usepackage{capt-of}
\usepackage{algorithm}
\usepackage{algpseudocode}
\usepackage{algorithmicx}
\usepackage{wasysym}
\usepackage[most]{tcolorbox}
\usepackage{amsmath}

\usepackage{colortbl}
\usepackage{pifont}
\usepackage{amssymb}
\usepackage{ulem} 
\usepackage{adjustbox}
\usepackage{graphicx}
\usepackage[utf8]{inputenc} 
\usepackage{makecell}
\usepackage{bbm}
\usepackage{multirow}
\usepackage{mathtools}
\usepackage{enumitem}
\usepackage{adjustbox}
\usepackage{setspace}
\usepackage{ulem}
\usepackage{capt-of}
\usepackage{algorithm}
\usepackage{algpseudocode}
\usepackage{algorithmicx}
\usepackage{hyperref}
\usepackage{url}
\usepackage[most]{tcolorbox}
\usepackage{amssymb}
\usepackage{arydshln}
\usepackage{graphicx}
\usepackage{hyperref}
\usepackage{url}
\usepackage{graphicx}
\usepackage{subcaption} 
\usepackage{caption}     
\usepackage{booktabs} 
\usepackage[most]{tcolorbox}
\usepackage{wrapfig}
\usepackage{enumitem}
\usepackage{multirow}
\usepackage{authblk}
\usepackage{listings}

\newcommand{\cmark}{\ding{51}}
\newcommand{\xmark}{\ding{55}}
\usepackage[table]{xcolor}

\definecolor{headerblue}{HTML}{ECF4FF}
\definecolor{deltagreen}{HTML}{4CAF50} 
\definecolor{headergray}{gray}{0.90}  
\definecolor{modelgray}{gray}{0.96}

\title{LeRF: Learning Reference Coordinate Frames for Perspective Taking Reasoning}

\usepackage{listings}
\author{
\textbf{Bang Xiao}$^{1,2}$ \quad
\textbf{Wenqi Jia}$^{1}$ \quad
\textbf{Ozgur Kara}$^{1}$ \quad
\textbf{Tiancheng Shen}$^{3}$ \\
\textbf{Yibo Yang}$^{4}$ \quad
\textbf{Bolin Lai}$^{1,5}$%
\addtocounter{footnote}{1}%
\thanks{Corresponding authors.} \quad
\textbf{Junho Kim}$^{1}$\protect\footnotemark[2] \quad
\textbf{James Matthew Rehg}$^{1}$\protect\footnotemark[2] \\ \vspace{2mm}
$^1$ University of Illinois Urbana-Champaign \quad
$^2$ Zhiyuan College, Shanghai Jiao Tong University \\
$^3$ University of California, Merced \quad
$^4$ Shanghai Jiao Tong University \quad
$^5$ Amazon AGI \\
}

\begin{document}
\maketitle
\begin{center}
\href{https://lerf-project.github.io/}{\raisebox{-0.15em}{
        \includegraphics[height=1.2em]{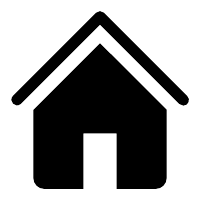}
    }
    Homepage}
\quad
\href{https://huggingface.co/collections/SamuelBang/lerf}{
    \raisebox{-0.15em}{
        \includegraphics[height=1.2em]{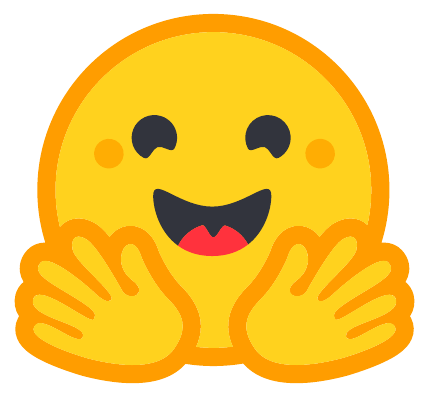}
    }
    Models
}
\end{center}

\begin{abstract}
Perspective taking is a fundamental component of spatial intelligence, requiring models  interpret spatial relations from a specified viewpoint, such as that of another entity or an imagined observer. Despite the increasing spatial reasoning capabilities of Vision-Language Models (VLMs), they still struggle with perspective taking, often defaulting to the camera viewpoint when a query requires reasoning from a different perspective. We introduce \underline{\textbf{Le}}arning \underline{\textbf{R}}eference Coordinate \underline{\textbf{F}}rames for Perspective Taking (\textbf{LeRF}), a framework that trains VLMs to construct and use explicit reference frames for viewpoint-dependent reasoning. Given an image and a query, LeRF decides whether a coordinate frame is necessary. If so, it grounds the reference entity and predicts the frame's origin and entity-centered reference frame. A lightweight renderer overlays the frame onto the image, enabling subsequent reasoning over these visual cues without external perception models or explicit 3D reconstruction. To learn this process, we first perform supervised fine-tuning to teach selective tool invocation and reference coordinate frame prediction, followed by reinforcement learning on spatial VQA pairs to improve frame-guided reasoning. Across diverse perspective-taking benchmarks, LeRF consistently improves over its backbone and achieves strong performance against existing open-source methods. Further evaluations also show improved reference-frame grounding and orientation estimation, supporting the effectiveness of learned reference frames for viewpoint-dependent reasoning.
\end{abstract}

\begin{figure*}[t]
    \centering
    \includegraphics[width=\textwidth]{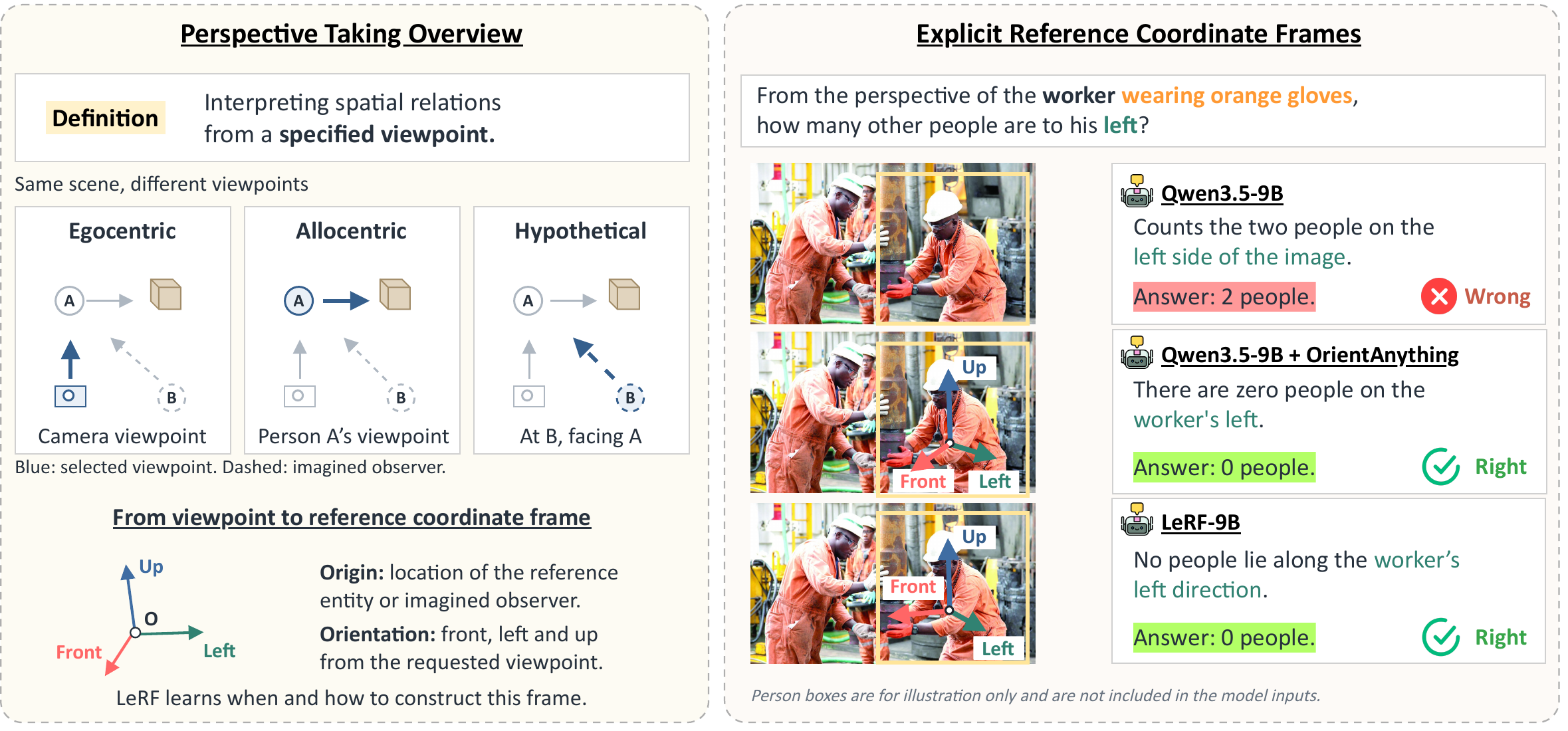}
    \caption{
    \textbf{Motivation of LeRF.} Perspective taking requires reasoning with respect to the viewpoint specified by a spatial query. An explicit entity-centered reference frame makes this viewpoint visually accessible, helping VLMs resolve viewpoint-dependent spatial relations and motivating LeRF to learn and reason with self-grounded reference frames.
    }
    \label{fig:teaser}
\end{figure*}


\section{Introduction}
Perspective taking~\citep{yu2017transformations,michelon2006two} requires interpreting spatial relations from viewpoints beyond the camera, including viewpoints of another entity or an imagined observer (as shown in Fig. \ref{fig:teaser}). This capability is a fundamental component of spatial intelligence in robotics and embodied AI domains \citep{huang2024rekepspatiotemporalreasoningrelational,du2024embspatial,lee2026multihopspatialmultihopcompositionalspatial}, especially for agent navigation, social interaction, collaboration, etc.
Due to the intrinsic difficulty in perspective transformation, frontier Vision-Language Models (VLMs) still struggle to distinguish the requested perspective from the camera viewpoint, leading to incorrect spatial judgments \citep{jia2026omnispatialcomprehensivespatialreasoning,li2025viewspatialbenchevaluatingmultiperspectivespatial}.
A central challenge is therefore to correctly identify the requested viewpoint and consistently interpret the scene from that perspective, rather than defaulting to the camera view.

Recent studies such as SpatialReasoner \citep{ma2025spatialreasonerexplicitgeneralizable3d} improve spatial reasoning by introducing explicit 3D representations shared across perception, computation, and reasoning stages. However, this design places a substantial burden on accurate 3D perception: errors in the inferred 3D representation directly propagate to downstream reasoning, making perception a key bottleneck of the framework.
Another line of work~\citep{lee2025perspectiveawarereasoningvisionlanguagemodels} adopts an agentic pipeline that reconstructs a coarse 3D scene with external vision models, transforms it into the target viewpoint, and feeds the transformed representation to the VLM. However, its multi-stage design incurs additional inference overhead and assumes perspective transformation is always required, limiting flexibility in more general spatial reasoning settings.

The key information for perspective taking extends beyond the location of the relevant observer or entity. It also includes how spatial directions are oriented from that viewpoint.
This naturally induces an entity-centered reference frame, grounded at the reference entity and oriented along its intrinsic front, left, and up directions. 
Making such a frame explicit in the image could turn an implicit viewpoint transformation into visually grounded directional reasoning \citep{https://doi.org/10.1111/tops.12233}. 
We first test this idea using OrientAnything-V2~\citep{wang2026orientv2unifyingorientation} to estimate the orientation of the reference entity and overlay the corresponding projected frame on the input image. 
Remarkably, this simple modification substantially improves perspective-taking accuracy (in Fig. \ref{fig:teaser}), suggesting that an explicit reference frame provides a useful intermediate representation for reasoning from another viewpoint. 
This observation motivates our central question: \textit{can a VLM learn to ground and establish its own reference coordinate frame, and use it to reason from the requested perspective?}

In this paper, we introduce \underline{\textbf{Le}}arning \underline{\textbf{R}}eference Coordinate \underline{\textbf{F}}rames for Perspective Taking (\textbf{LeRF}), a framework that enables a VLM to autonomously establish, externalize, and reason over its own reference frames.
Rather than relying on external 3D perception models or struggling with implicit viewpoint shifts, LeRF learns to dynamically externalize the requested perspective: it grounds the reference entity, instantiates its projected directional frame directly onto the visual scene, and revisits this self-generated visual scaffold to guide downstream reasoning.
In doing so, LeRF transforms an otherwise challenging mental rotation into grounded visual perception. Additionally, LeRF can also adaptively decides whether to predict and render the reference coordinate frame. It avoids unnecessary tool call when there is no perspective transformation.

To learn when to construct reference frames and how to use them for reasoning, we train LeRF through supervised fine-tuning (SFT) followed by reinforcement learning (RL).
We first construct reference-frame data from public object and human pose datasets \citep{black2023bedlamsyntheticdatasetbodies,ma2024imagenet3dgeneralpurposeobjectlevel3d,zhang2024omni6dposebenchmarkmodeluniversal}.
SFT teaches reference-entity grounding and projected frame prediction, together with selective tool use through examples that do not require frame construction.
We then apply RL on public perspective-taking VQA datasets \citep{lee2026multihopspatialmultihopcompositionalspatial,ma2025spatialreasonerexplicitgeneralizable3d}, using final-answer correctness as the reward to consolidate reasoning with self-predicted coordinate frames.

Extensive experiments across diverse perspective-taking benchmarks~\citep{jia2026omnispatialcomprehensivespatialreasoning, ma20253dsrbenchcomprehensive3dspatial, li2025viewspatialbenchevaluatingmultiperspectivespatial} 
demonstrate that LeRF consistently outperforms existing open-source VLMs, achieving state-of-the-art performance on non-camera viewpoints while competing favorably with specialized pose estimators.
Crucially, our analyses confirm that the model actively leverages these externalized directional cues: systematic frame perturbations induce clear performance drops, and explicit visual rendering provides an indispensable anchor especially for hypothetical viewpoints.
Furthermore, ablations show that frame-supervised initialization provides useful spatial knowledge for subsequent reasoning, even when explicit overlays are removed.

Our contributions are threefold:
\begin{itemize}[leftmargin=1.2em, itemsep=0.2em, topsep=0.2em]
    \item We demonstrate that projected entity-centered reference frames provide effective visual scaffolding for perspective taking, enabling directional reasoning directly on the image without explicit 3D reconstruction.
    \item We develop LeRF, which empowers a VLM to selectively ground, establish and reason with its own reference frames via supervised frame prediction and reinforcement learning, relying only on a deterministic renderer at inference time.
    \item We establish new state-of-the-art results on multiple perspective-taking benchmarks among open-source VLMs while rivaling specialized pose estimators, and provide in-depth diagnostic analyses demonstrating the causal role of visual rendering and the internal spatial prior instilled by frame-prediction training.
\end{itemize}

\section{Related Works}

\paragraph{Perspective-Taking Reasoning in VLMs.}
Perspective taking requires reasoning about spatial relations under a reference frame that may differ from the camera viewpoint. Recent studies consistently reveal substantial limitations of VLMs in this capability. Benchmarks such as COMFORT \citep{zhang2025visionlanguagemodelsrepresentspace}, ViewSpatial-Bench \citep{li2025viewspatialbenchevaluatingmultiperspectivespatial}, SpinBench \citep{zhang2026spinbenchperspectiverotationlens}, and OmniSpatial \citep{jia2026omnispatialcomprehensivespatialreasoning} show that current models exhibit strong camera-centric biases, limited robustness to changes in spatial frames of reference, and significant performance degradation under viewpoint transformations. Current methods such as SpatialReasoner \citep{ma2025spatialreasonerexplicitgeneralizable3d} explicitly estimate 3D point coordinates and derive relative spatial directions through geometric computations, but remain constrained by errors in the VLM's 3D perception and coordinate estimation. APC \citep{lee2025perspectiveawarereasoningvisionlanguagemodels}, in contrast, constructs a coarse 3D scene abstraction using multiple external perception models and transforms it into the reference entity's egocentric frame, introducing additional system complexity and potential error propagation across intermediate modules.

\paragraph{Structured Spatial Representations for VLM Reasoning.}
A growing line of work improves spatial reasoning by introducing structured intermediate representations that make geometric information more accessible to VLMs.
SpatialReasoner \citep{ma2025spatialreasonerexplicitgeneralizable3d} explicitly represents scene geometry with 3D coordinates shared across perception, computation, and reasoning, while Thinking with Blueprints \citep{ma2026thinkingblueprintsassistingvisionlanguage} constructs an object-centric structured representation containing the positions, sizes, and attributes of relevant entities before reasoning over it.
These studies demonstrate the value of introducing geometric structure into the reasoning process.
Our work focuses on a lightweight object-centered reference frame tailored to perspective transformation, which can be directly grounded and visualized in the original image without reconstructing the full 3D scene.

\section{LeRF}

\begin{figure*}[t]
    \centering
    \includegraphics[width=\textwidth]{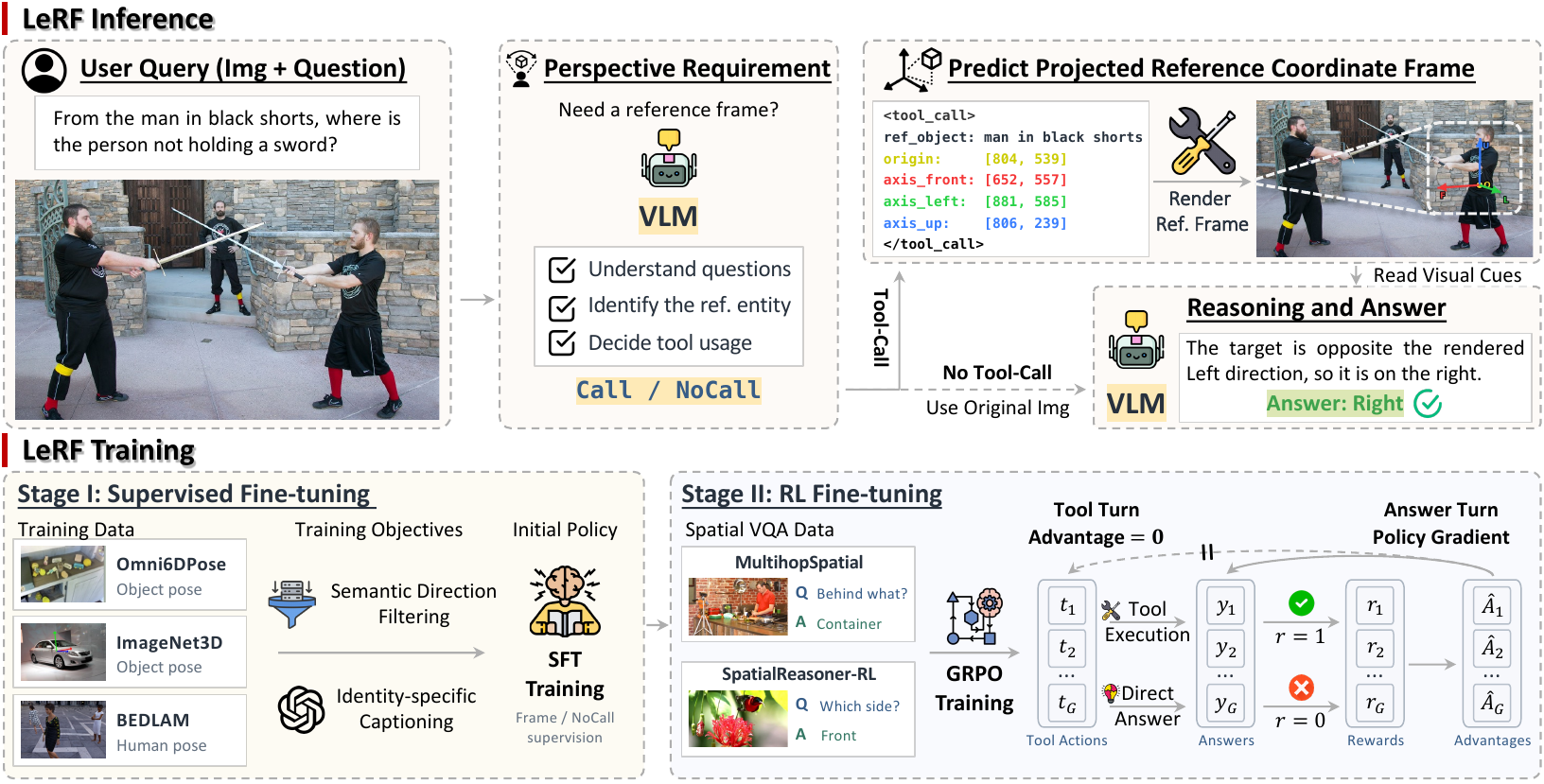}
    \caption{\textbf{Overview of LeRF.} The VLM selectively constructs and renders a reference frame for perspective-taking reasoning, or directly reasons from the original image when no frame is needed. LeRF is trained with SFT for frame prediction and RL for reasoning with self-predicted frames.}
    \label{fig:pipeline}
\end{figure*}

Inspired by human's mind in perspective taking, LeRF consists of a self-grounded reference-frame construction and reasoning process illustrated in Fig.~\ref{fig:pipeline}. The core idea is to transform abstract viewpoint shifts into concrete visual evidence: the model explicitly predicts reference axes, render the reference axes onto the image via tool calling, and grounds subsequent reasoning on rendered cues.

\subsection{Reference Coordinate Frame Formulation and Rendering}
\label{sec:method-reference_frame}

\paragraph{Reference Entity Grounding and Frame Prediction.}
Given an input image $\mathbf{I}$ and a spatial query $q$, LeRF first identifies the reference entity specified or implied by the query and grounds it in the image. It then predicts an object-centered reference frame attached to the grounded entity, where the $+X$, $+Y$, and $+Z$ axes respectively correspond to its intrinsic front, left, and up directions.

Different from solutions recovering the reference entity in a full 3D coordinate system, LeRF predicts the projection of its object-centered frame onto the 2D image plane. This design is motivated by two considerations. First, a projected reference frame already provides the directional cues required for most perspective-taking and spatial reasoning problems, analogous to how humans reason about 3D configurations using 2D geometric sketches and projected coordinate frames \citep{yu2017transformations,schultheis2022smart}. Second, recovering accurate metric 3D coordinates from a single image remains intrinsically challenging and can introduce substantial estimation errors, which may subsequently propagate to downstream spatial reasoning \citep{ma2025spatialreasonerexplicitgeneralizable3d}. Therefore, adopting a simple 2D representation can clearly indicate the entity's orientation, and avoid compounding mistakes from 3D reconstruction in the meanwhile.

Formally, the predicted reference frame is represented as
\begin{equation}
\widehat{\mathcal{F}}
=
\left\{
\hat{\mathbf{o}},
\hat{\mathbf{p}}_{\mathrm{front}},
\hat{\mathbf{p}}_{\mathrm{left}},
\hat{\mathbf{p}}_{\mathrm{up}}
\right\} \quad \hat{\mathbf{o}}, \hat{\mathbf{p}}_\mathrm{front}, \hat{\mathbf{p}}_\mathrm{left}, \hat{\mathbf{p}}_\mathrm{up}\in\mathbb{R}^2,
\end{equation}
where $\hat{\mathbf{o}}$ denotes the predicted projected origin, and
$\hat{\mathbf{p}}_{\mathrm{front}}$,
$\hat{\mathbf{p}}_{\mathrm{left}}$, and
$\hat{\mathbf{p}}_{\mathrm{up}}$ denote the predicted projected endpoints of
the three axes. For an entity-centered viewpoint, the origin is anchored to
the reference entity, with the axes aligned to its intrinsic front, left, and
up directions. For a hypothetical viewpoint, the origin and axes instead
reflect the imagined observer's location and orientation specified or implied
by the query. To align with the representation used in Qwen3.5
\citep{qwen3.5} pretraining, we represent each point using integer
coordinates in $[0,1000]^2$, with the top-left corner of the image as the
origin. This representation jointly captures reference-entity grounding and
orientation, and can be directly rendered onto the original image as an
explicit visual cue for subsequent perspective-taking reasoning.

\paragraph{Reference Coordinate Frame Rendering.}
After predicting $\widehat{\mathcal{F}}$, LeRF invokes a lightweight,
deterministic rendering function $\mathcal{R}$ that converts the predicted
frame parameters into a visual overlay. Such a 2D visual representation provides visible directional information needed while avoiding the complexity and potential error propagation of explicit 3D reconstruction. Specifically, the model passes the
predicted origin and the endpoints of the front, left, and up axes to the
renderer, which overlays the corresponding reference frame onto the original image (see Fig. \ref{fig:pipeline} top-right example):
\begin{equation}
\widetilde{\mathbf{I}}
=
\mathcal{R}\left(\mathbf{I},\widehat{\mathcal{F}}\right).
\end{equation}
For axes nearly perpendicular to the image plane and therefore
collapse to a very short 2D projection, we explicitly encode their depth
direction using standard geometric symbols: $\odot$ denotes that the axis points toward the camera and out of the image plane, while $\otimes$
denotes that it points away from the camera and into the image plane. This
avoids losing orientation information when perspective projection makes an
axis visually degenerate. Details are listed in Appendix \ref{sec:app-oop-render}.

\subsection{Pipeline: Reasoning with Reference Coordinate Frames}
\label{sec:method-frame_reasoning}

LeRF performs perspective-taking through a two-stage inference process as shown in Fig. \ref{fig:pipeline}. Given
an image $\mathbf{I}$ and a spatial query $q$, the model first generates an
intermediate action
\begin{equation}
a_1 \sim \pi_\theta(\cdot \mid q,\mathbf{I}), \quad \text{where}\; a_1
\in
\left\{
\mathrm{\texttt{NoCall}},
\mathrm{\texttt{Call}}\left(\hat e,\widehat{\mathcal F}\right)
\right\}.
\end{equation}
Here, $\hat e$ denotes the predicted reference entity and
$\widehat{\mathcal{F}}$ denotes its predicted reference frame. \texttt{Call} and \texttt{NoCall} denote tool-call and
no-tool-call actions, respectively. It is worth noting that the model has to make a decision whether to call the tool based on the question. The \texttt{NoCall} action allows the model to bypass explicit frame
construction when perspective transformation is unnecessary (e.g., from the egocentric view). The visual output returned by the tool is
\begin{equation}
v_1
=
\begin{cases}
\mathcal{R}\left(\mathbf I,\widehat{\mathcal F}\right),
& \text{if } a_1=\mathrm{\texttt{Call}}\left(\hat e,\widehat{\mathcal F}\right),\\
\varnothing,
& \text{if } a_1=\mathrm{\texttt{NoCall}}.
\end{cases}
\end{equation}

Before the second inference stage, we remove the numerical reference-frame
parameters from the first-stage action $a_1$ and keep only the rendered image with predicted XYZ-axes, resulting in $\bar a_1$. 
The subsequent reasoning and response are grounded on the visual directional signals and all the previous contexts:
\begin{equation}
y\sim\pi_\theta(\cdot \mid q,\mathbf{I},\bar a_1,v_1).
\end{equation}
Thus, the model retains its first-stage tool-use decision (whether to call tools) and the predicted reference entity, but have no access to the
numerical parameters of the predicted frame. When the tool is invoked, the
model needs to interpret the visually rendered frame in $v_1$ and combine
its directional cues with spatial evidence in the original image to
reason from the requested perspective. 

\subsection{Model Training}
\label{sec:method-training}

\paragraph{Stage I: Supervised Fine-Tuning (SFT).}
The first stage teaches the model to identify the reference entity and construct its reference frame from the input image and question. We organize the reference entity, the origin, and the endpoints of the three axes in a structured tool-call format as the training target. For questions that do not require perspective transformation, we include examples without reference-frame construction to teach the model when to invoke the tool or not. We optimize the model with a standard autoregressive cross-entropy loss over the target response, while keeping the vision encoder and multimodal projector frozen. This stage equips the model with reference entity grounding and pose prediction capabilities for subsequent reference-frame-based reasoning.

\paragraph{Stage II: Reinforcement Learning (RL).}
Building on the reference entity grounding and pose prediction capabilities acquired in Stage I, we further train the model for perspective-taking reasoning using Group Relative Policy Optimization (GRPO)~\citep{shao2024deepseekmathpushinglimitsmathematical}. Following VTool-R1~\citep{wu2026vtoolr1vlmslearnthink}, we adopt a binary reward based solely on final-answer correctness, assigning $1$ to correct answers and $0$ otherwise, without additional rewards for intermediate predictions or tool use. To avoid degrading the reference coordinate frame prediction capabilities learned during SFT, we exclude reference-frame prediction tokens from the policy-gradient objective and apply the advantage-weighted loss only to the subsequent reasoning and answer tokens (see Fig. \ref{fig:pipeline} Stage II).

During the reasoning turn, we mask the numerical frame coordinates in the tool-call history to prevent direct access to the predicted coordinates. Instead, the predicted frame is presented through the tool-rendered image, encouraging the model to interpret its visual directional cues and combine them with spatial evidence from the scene. In this stage, the model learns to apply its spatial orientation knowledge (learned in Stage I) to downstream perspective-taking tasks.

\subsection{Training Data}
\label{sec:training-data}
We curate training data to teach the model when to construct a reference frame, how to predict the reference entity's pose, and how to use its predicted frame for perspective-taking reasoning. For training stage I (SFT), we derive reference-frame supervision from object and human pose datasets and apply additional data processing. For training stage II (RL), we train on perspective-taking VQA datasets to train the model to apply these capabilities to spatial question answering.

\paragraph{Reference Frame Prediction Data.}
We derive reference-frame labels from the object and human poses provided by
ImageNet3D~\citep{ma2024imagenet3dgeneralpurposeobjectlevel3d},
Omni6DPose-SOPE~\citep{zhang2024omni6dposebenchmarkmodeluniversal}, and
BEDLAM~\citep{black2023bedlamsyntheticdatasetbodies}.
We align each dataset's canonical frame with our front, left, and up axes,
manually specifying category-level mappings for Omni6DPose-SOPE and
excluding categories without well-defined orientations
(Appendix~\ref{sec:filtered-obj-category-for-sft}).
Given an entity's camera-frame pose $(\mathbf{R},\mathbf{t})$ and
camera intrinsics $\mathbf{K}$, we obtain the frame through
first-order perspective projection:
\begin{equation}
\mathbf{o}=\pi_{\mathbf{K}}(\mathbf{t}),
\qquad
\mathbf{p}_d=\mathbf{o}
+s\,\mathbf{J}_{\pi_{\mathbf{K}}}(\mathbf{t})\,\mathbf{R}\mathbf{e}_d,
\end{equation}
where $\pi_{\mathbf{K}}$ denotes pinhole projection,
$\mathbf{J}_{\pi_{\mathbf{K}}}$ its Jacobian, and $\mathbf{e}_d$ the
canonical unit axis for $d\in\{\mathrm{front},\mathrm{left},\mathrm{up}\}$.
A shared scale $s$ preserves relative projected axis lengths,
and coordinates are normalized to integers in $[0,1000]^2$.

We use GPT-5.6-Terra~\citep{openai2026gpt56} to generate appearance-based
referring descriptions for BEDLAM (verified through grounding), and to
filter ambiguous multi-instance samples from Omni6DPose-SOPE.
Each entity is paired with a template-based orientation or
perspective-taking question, with its frame encoded as a native tool call
(Appendix~\ref{sec:qtemplate-tool-calling}). System prompt and further data construction details are provided in Appendix~\ref{appendix:frame-label-construction}, \ref{appendix:system_prompt} respectively.

\paragraph{No-Tool-Call Supervision.}
Not every spatial question requires constructing an entity-centered reference frame; indiscriminate tool invocation would introduce unnecessary visual transformations and computation. We therefore include no-tool-call data to teach the model to distinguish questions that require perspective transformation from those that can be answered directly from the original image. Specifically, we reuse selected images to construct questions involving camera-relative relations, image positions and counting, constituting approximately $20\%$ of the final SFT dataset.. The model learns to output a predefined special token, \texttt{NO\_TOOL\_CALL}, in these training samples. Detailed prompt templates are listed in Appendix~\ref{sec:qtemplate-no-tool-calling}.

\paragraph{Perspective-Taking Reasoning Data.}
We curate two open-sourced perspective taking VQA datasets for RL training: (1) MultihopSpatial-train \citep{lee2026multihopspatialmultihopcompositionalspatial} which contains $6.79$K perspective taking queries with 1 to 3 reasoning hops and (2) SpatialReasoner-RL \citep{ma2025spatialreasonerexplicitgeneralizable3d} which contains $1.2$K perspective taking QA pairs. 
\section{Experiments}
\label{sec:exp}

\paragraph{Implementation Details.} We implement LeRF on two models with different parameter scales: Qwen3.5-4B and Qwen3.5-9B \citep{qwen3.5}. In SFT training (Stage I), we adopt LoRA \citep{hu2021loralowrankadaptationlarge} with a rank of $8$ using LLaMA-Factory \citep{zheng2024llamafactory}. In RL training (Stage II), we adopt GRPO \citep{shao2024deepseekmathpushinglimitsmathematical} using VeRL \citep{sheng2024hybridflow}. For both SFT and RL, we keep the vision encoder and multimodal projector frozen and update only the parameters of the LLM backbone. All trainings are conducted in $4\times$NVIDIA RTX PRO 6000 Blackwell GPUs. Details are provided in Appendix \ref{sec:training-details}.

\begin{table*}[t]
\centering
\caption{
Quantitative results on OmniSpatial-Perspective Taking
\citep{jia2026omnispatialcomprehensivespatialreasoning} (OmniSpatial-PT),
3DSRBench \citep{ma20253dsrbenchcomprehensive3dspatial},
and ViewSpatial-Bench (V-Spatial-Bench)
\citep{li2025viewspatialbenchevaluatingmultiperspectivespatial}.
Ori / M-Obj denote Orientation and Multi-Object, and P-Obj / P-Rel denote
Person Perspective--Object View Orientation and
Person Perspective--Relative Direction, respectively.
Except for SpatialReasoner, which does not support thinking,
all models enable thinking during evaluation.
The best and second-best results among open-source methods are shown in
\textbf{bold} and \underline{underlined}.
}
\label{tab:main_results}

\setlength{\tabcolsep}{5pt}

\resizebox{0.85\linewidth}{!}{
\begin{tabular}{
    l!{\vrule width 0.4pt}
    ccc!{\vrule width 0.4pt}
    cc!{\vrule width 0.4pt}
    cc
}
\toprule

\multicolumn{1}{c!{\vrule width 0.4pt}}
{\multirow{2}{*}[-0.5ex]{Method}}
&
\multicolumn{3}{c!{\vrule width 0.4pt}}{OmniSpatial-PT}
&
\multicolumn{2}{c!{\vrule width 0.4pt}}{3DSRBench}
&
\multicolumn{2}{c}{ViewSpatial-Bench}
\\

\cmidrule(lr){2-4}
\cmidrule(lr){5-6}
\cmidrule(lr){7-8}

& Ego & Allo & Hypo
& Ori & M-Obj
& P-Obj & P-Rel
\\

\Xhline{2\arrayrulewidth}

\multicolumn{8}{l}{
    \cellcolor{headergray}\textit{Proprietary Model}
}
\\

\textcolor{gray}{GPT-5.6-Luna (medium)}
& \textcolor{gray}{83.33}
& \textcolor{gray}{49.73}
& \textcolor{gray}{45.78}
& \textcolor{gray}{60.04}
& \textcolor{gray}{55.34}
& \textcolor{gray}{46.99}
& \textcolor{gray}{70.07}
\\

\textcolor{gray}{GPT-5.6-Terra (medium)}
& \textcolor{gray}{81.37}
& \textcolor{gray}{55.85}
& \textcolor{gray}{53.01}
& \textcolor{gray}{63.32}
& \textcolor{gray}{56.39}
& \textcolor{gray}{45.08}
& \textcolor{gray}{77.20}
\\

\textcolor{gray}{Claude Sonnet 5 (medium)}
& \textcolor{gray}{80.39}
& \textcolor{gray}{42.55}
& \textcolor{gray}{49.40}
& \textcolor{gray}{34.94}
& \textcolor{gray}{44.77}
& \textcolor{gray}{51.31}
& \textcolor{gray}{51.43}
\\

\textcolor{gray}{Claude Sonnet 5 (high)}
& \textcolor{gray}{84.31}
& \textcolor{gray}{48.14}
& \textcolor{gray}{45.78}
& \textcolor{gray}{43.15}
& \textcolor{gray}{46.81}
& \textcolor{gray}{51.51}
& \textcolor{gray}{60.10}
\\

\hline

\multicolumn{8}{l}{
    \cellcolor{headergray}\textit{Open-sourced Generalist}
}
\\

InternVL3.5-4B
& 67.25 & 31.54 & 41.45
& 30.31 & 39.26
& 48.29 & 40.02
\\

InternVL3.5-8B
& 67.84 & 34.20 & 40.24
& 27.70 & 38.33
& 57.63 & 42.40
\\

GLM-4.6V-Flash
& 74.31 & 34.52 & 39.52
& 38.32 & 44.13
& 60.36 & 43.94
\\

Qwen3.5-4B
& \underline{74.71}
& 42.55
& 44.34
& 42.28
& 43.26
& 51.01
& 57.43
\\

Qwen3.5-9B
& \textbf{80.20}
& 47.13
& 44.58
& 48.17
& 48.66
& 56.23
& 65.51
\\

\hline

\multicolumn{8}{l}{
    \cellcolor{headergray}\textit{Open-sourced Specialist}
}
\\

Qwen3.5-4B + APC
& 41.18
& 27.39
& 34.94
& 42.66
& 31.18
& \underline{60.76}
& 37.41
\\

Qwen3.5-9B + APC
& 42.16
& 27.66
& 30.12
& 44.98
& 33.22
& 59.34
& 37.53
\\

SpatialReasoner
& 40.39
& 35.11
& 35.66
& \underline{52.05}
& \textbf{50.64}
& 42.37
& 45.61
\\

\hline

\multicolumn{8}{l}{
    \cellcolor{headerblue}\textit{Our Framework}
}
\\

LeRF-4B
& 72.35
& \underline{49.36}
& \underline{46.75}
& 45.88
& 44.55
& 56.26
& \underline{67.85}
\\

LeRF-9B
& 74.31
& \textbf{54.04}
& \textbf{55.66}
& \textbf{53.76}
& \underline{50.29}
& \textbf{61.91}
& \textbf{74.23}
\\

\bottomrule
\end{tabular}
}

\end{table*}

\paragraph{Experimental Setting.}
We evaluate LeRF on three public benchmarks.
For OmniSpatial~\citep{jia2026omnispatialcomprehensivespatialreasoning}, we use its perspective-taking subset (OmniSpatial-PT), which tests spatial reasoning from camera (Ego), entity-centered (Allo), and imagined (Hypo) viewpoints.
For 3DSRBench~\citep{ma20253dsrbenchcomprehensive3dspatial}, we evaluate the Orientation and Multi-Object categories, covering object orientations and spatial relations among multiple objects.
For ViewSpatial-Bench~\citep{li2025viewspatialbenchevaluatingmultiperspectivespatial}, we select the person-perspective Object View Orientation and Relative Direction tasks, which assess facing directions and object locations from a person's viewpoint.
We compare against three groups of baselines:
(1) proprietary models, including GPT-5.6~\citep{openai2026gpt56} and Claude Sonnet 5~\citep{anthropic2026claudesonnet5};
(2) open-source generalists, including Qwen3.5~\citep{qwen3.5}, GLM-4.6V-Flash~\citep{vteam2026glm45vglm41vthinkingversatilemultimodal}, and InternVL3.5~\citep{wang2025internvl35advancingopensourcemultimodal};
and (3) specialist methods, including SpatialReasoner~\citep{ma2025spatialreasonerexplicitgeneralizable3d} and APC~\citep{lee2025perspectiveawarereasoningvisionlanguagemodels}.

\subsection{Performance on Perspective-Taking Benchmarks}
As shown in Table~\ref{tab:main_results}, LeRF consistently improves over its corresponding Qwen3.5 backbone across the evaluated settings which need viewpoint changing.
LeRF-9B achieves the best results among open-source methods on OmniSpatial-PT Allo and Hypo, 3DSR Orientation, and both ViewSpatial-Bench categories, while ranking second on 3DSR Multi-Object.
Compared with Qwen3.5-9B, it improves accuracy by $6.91$ percentage points (pp) and $11.08$ pp on Allo and Hypo respectively, and by $8.72$ pp on ViewSpatial-Bench P-Rel set, demonstrating substantial gains in reasoning from another entity's or an imagined viewpoint.
LeRF-4B also consistently outperforms its backbone on these perspective-taking tasks, indicating that the benefits extend across model sizes.
Notably, LeRF-9B surpasses all evaluated proprietary models on OmniSpatial-PT Hypo and ViewSpatial-Bench P-Obj set.
Accuracy on OmniSpatial-PT Ego decreases relative to the backbones, indicating a trade-off between improved perspective-taking performance and performance on camera-centric questions.
Overall, these results support the effectiveness of learning to construct and reason with reference frames for perspective taking. More comparisons are listed in Appendix \ref{appendix:latency}, \ref{appendix:more_case_study}.

Beyond perspective-taking performance, we directly evaluate whether LeRF learns to accurately ground and orient the reference frames used for reasoning.
We evaluate projected reference-frame estimation on sampled subsets of EMDB~\citep{kaufmann2023emdbelectromagneticdatabaseglobal} and OmniNOCS-Objectron~\citep{krishnan2024omninocsunifiednocsdataset,ahmadyan2020objectronlargescaledataset}, containing 7,273 human views and 954 object views, respectively. All models receive full images. LeRF and Qwen3.5 predict the frame origin and three projected axis endpoints, while OrientAnything-V2's~\citep{wang2026orientv2unifyingorientation} predicted 3D orientations are projected using ground-truth camera intrinsics and origins. We report All@$\theta$, requiring all three projected axis-direction errors to be within $\theta$, and Origin@0.1Diag, requiring the origin error to be within $10\%$ of the ground-truth bounding-box diagonal.

\subsection{Reference Frame Estimation Accuracy}
\begin{table*}[t]
\centering
\caption{
Projected reference-frame accuracy (\%) on sampled EMDB and
OmniNOCS-Objectron subsets.
All@$\theta$ requires all three 2D axis-direction errors to be
within $\theta$.
Origin@0.1Diag denotes the percentage of views whose predicted origin
is within 10\% of the ground-truth bounding-box diagonal from the
ground-truth origin.
Best and second-best results are \textbf{bold} and
\underline{underlined}.
}
\label{tab:pose_evaluation}

\setlength{\tabcolsep}{5pt}

\resizebox{0.9\linewidth}{!}{
\begin{tabular}{lcccccc}
\toprule
\multirow[c]{2}{*}[-0.5ex]{Model}
& \multicolumn{3}{c}{EMDB}
& \multicolumn{3}{c}{OmniNOCS-Objectron} \\
\cmidrule(lr){2-4} \cmidrule(lr){5-7}
& All@15$^\circ$ & All@30$^\circ$ & Ori@0.1Diag
& All@15$^\circ$ & All@30$^\circ$ & Ori@0.1Diag \\
\hline

\multicolumn{7}{l}{
    \cellcolor{headergray}\textit{Baseline VLMs}
} \\
Qwen3.5-4B
& 6.87 & 17.89 & 20.45
& 2.73 & 5.66 & 53.04 \\

Qwen3.5-9B
& 1.11 & 4.43 & 91.63
& 0.52 & 1.68 & 70.34 \\
\hline

\multicolumn{7}{l}{
    \cellcolor{headergray}\textit{Expert Model}
} \\
OriAnything-V2
& 19.35 & 44.84 & --
& \textbf{50.42} & \textbf{83.23} & -- \\
\hline

\multicolumn{7}{l}{
    \cellcolor{headerblue}\textit{Our Framework}
} \\
LeRF-4B
& \underline{36.15} & \underline{65.50} & \underline{99.81}
& 39.10 & 65.51 & \underline{81.55} \\

LeRF-9B
& \textbf{43.12} & \textbf{70.05} & \textbf{99.82}
& \underline{49.58} & \underline{72.54} & \textbf{87.95} \\
\bottomrule
\end{tabular}
}
\end{table*}

As presented in Table~\ref{tab:pose_evaluation}, both LeRF variants substantially improve reference-frame estimation over their backbones. LeRF-9B surpasses OrientAnything-V2 on EMDB by $23.77$ pp and $25.21$ pp at All@$15^\circ$ and All@$30^\circ$, respectively. On OmniNOCS-Objectron, it nearly matches OrientAnything-V2 at All@$15^\circ$ but trails at All@$30^\circ$. LeRF-9B also achieves the best origin localization accuracy on both datasets, reaching $99.82\%$ and $87.95\%$, respectively.

\subsection{Analysis and Ablation Studies}
\begin{figure}[t]
  \centering

  \begin{minipage}[t]{0.48\linewidth}
    \centering
    \includegraphics[width=.95\linewidth]{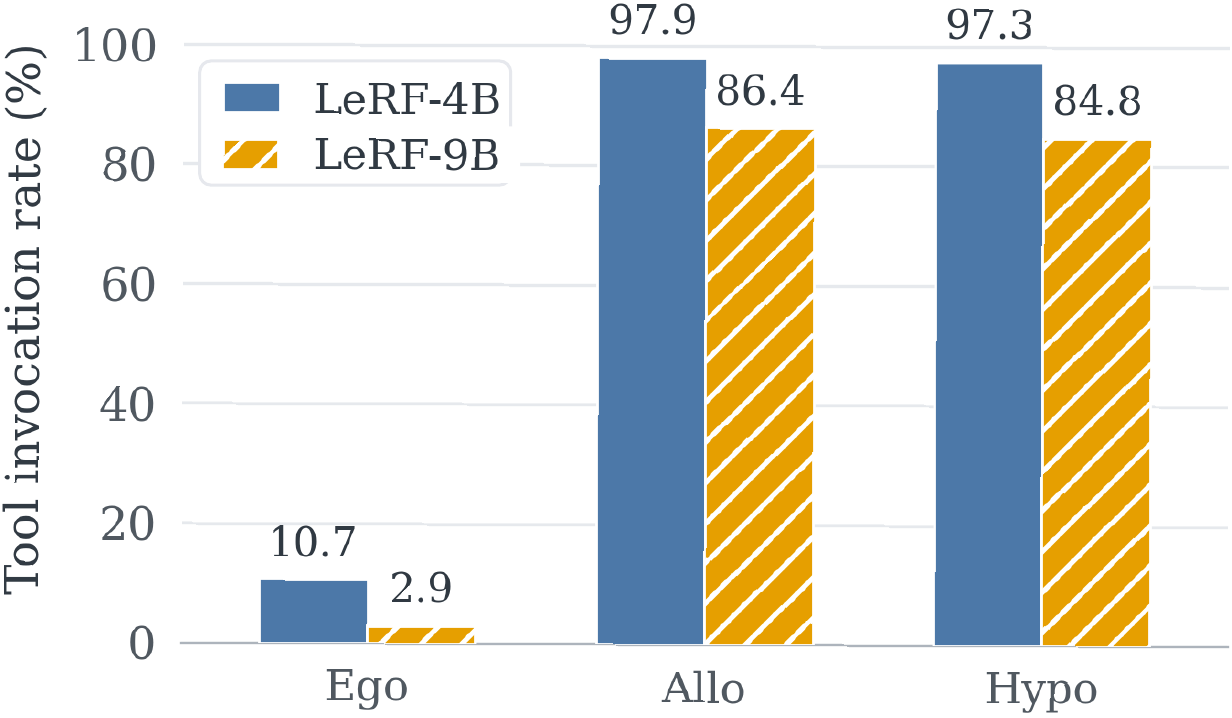}
    \captionof{figure}{
      Tool invocation rates (\%) across egocentric, allocentric, and hypothetical question types in OmniSpatial-PT.
    }
    \label{fig:4}
  \end{minipage}
  \hfill
  \begin{minipage}[t]{0.48\linewidth}
    \centering
    \includegraphics[width=.95\linewidth]{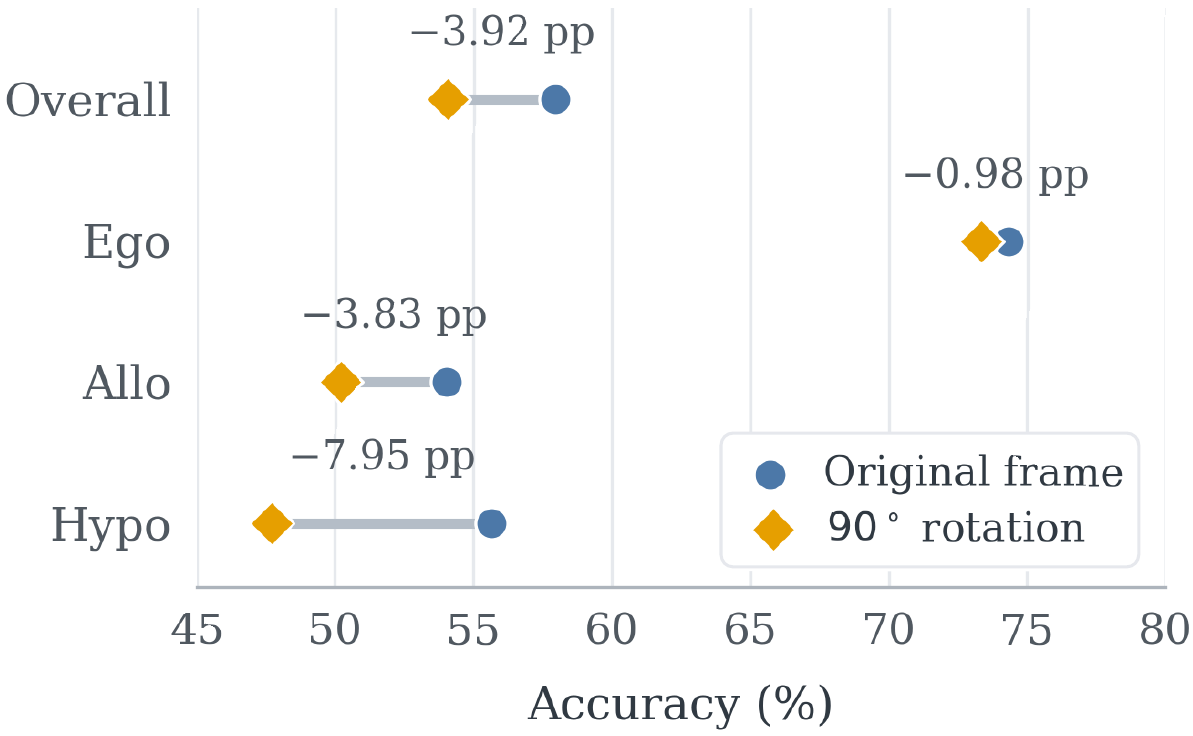}
    \captionof{figure}{
      Effect of perturbing the rendered reference frame
      at evaluation time on OmniSpatial-PT with LeRF-9B.
    }
    \label{fig:5}
  \end{minipage}
\end{figure}

\paragraph{Selective Tool Invocation.}
We analyze whether LeRF adapts tool use to the perspective required by
the question. As shown in Fig.~\ref{fig:4}, both models invoke the tool
substantially more often for allocentric and hypothetical questions,
which require another entity's or an imagined viewpoint, than for
egocentric questions.
LeRF-9B shows stronger selectivity, invoking the tool on only 2.94\%
of egocentric questions versus over 84\% for the other two categories,
suggesting that tool use aligns with the need for perspective transformation.

\paragraph{Use of Rendered Directional Cues.}

To test whether LeRF uses the rendered reference frame during inference, we evaluate
LeRF-9B on OmniSpatial-PT by rotating the frame $90^\circ$ counterclockwise
whenever the tool is invoked, keeping all other inference settings unchanged.
As shown in Fig.~\ref{fig:5}, overall accuracy drops from $57.97\%$ to
$54.05\%$, with the largest decline on Hypo (7.95 pp),
while Ego remains largely unaffected.
These results indicate that LeRF incorporates the rendered directional
cues into its reasoning rather than relying solely on internal orientation estimates.

\paragraph{Complementary Roles of SFT and RL.}
We compare the base model, the SFT-only checkpoint, and RL initialized
from either checkpoint, while varying rendering during RL and evaluation.
All RL variants undergo $50$ training steps; disabling rendering returns
the original image with the remaining setup unchanged.
As shown in Table~\ref{tab:rendering_ablation}, the SFT-only checkpoint achieves $40.86\%$ overall accuracy,
below the base model's $52.76\%$.
This drop may reflect degradation of the model's existing reasoning
capabilities during SFT.
Nevertheless, this checkpoint provides a useful initialization for RL:
SFT followed by RL achieves $57.18\%$ without rendering and $57.97\%$
with rendering during both RL and evaluation, compared with $54.34\%$
and $53.76\%$, respectively, for RL directly from the base model.
These results suggest complementary roles for the two stages:
SFT provides an initialization that subsequent RL can exploit more
effectively for spatial question answering, although SFT alone is
insufficient to improve task accuracy.

\paragraph{Role of Rendering During Training and Inference.}

\begin{table*}[t]
\centering
\caption{
Effect of SFT, RL, and reference-frame rendering on OmniSpatial-PT.
The base model and SFT-only checkpoint are included as baselines.
All RL variants are trained for $50$ steps.
When rendering is disabled, the tool returns the original image.
Accuracy is reported in percentages.
}
\label{tab:rendering_ablation}
\setlength{\tabcolsep}{7pt}

\resizebox{0.9\linewidth}{!}{
\begin{tabular}{lcccccc}
\toprule
RL Initialization & RL Render & Eval Render
& Ego & Allo & Hypo & Overall\\
\Xhline{2\arrayrulewidth}

\multicolumn{7}{l}{
    \cellcolor{headergray}\textit{Base Model}
} \\
Qwen3.5-9B (Base)
& -- & --
& 80.20 & 47.13 & 44.58 & 52.76 \\
\hline

\multicolumn{7}{l}{
    \cellcolor{headergray}\textit{SFT Baseline}
} \\
Qwen3.5-9B + SFT
& -- & \cmark
& 69.80 & 34.47 & 34.22 & 40.86 \\
\hline

\multicolumn{7}{l}{
    \cellcolor{headergray}\textit{Direct RL Fine-tuning (w/o SFT)}
} \\
\multirow{2}{*}{Qwen3.5-9B + RL}
& \xmark & \xmark
& 80.34 & 49.40 & 45.18 & 54.34 \\
& \cmark & \cmark
& 74.31 & 50.16 & 44.82 & 53.76 \\
\hline

\multicolumn{7}{l}{
    \cellcolor{headerblue}\textit{LeRF Training (Ours)}
} \\
\multirow{4}{*}{\makecell[l]{Qwen3.5-9B \\ \quad + SFT $\rightarrow$ RL}}
& \xmark & \xmark
& 72.94 & 54.63 & 49.40 & 57.18 \\
& \xmark & \cmark
& 69.80 & 53.78 & 55.42 & 56.93 \\
& \cmark & \xmark
& 76.47 & 50.37 & 50.60 & 55.15 \\
& \cmark & \cmark
& 74.31 & 54.04 & 55.66 & 57.97 \\
\bottomrule
\end{tabular}
}
\end{table*}

To isolate the role of explicit reference-frame visualization,
we vary rendering during RL and evaluation for models initialized
from the SFT checkpoint.
Training and evaluating without rendering achieves $57.18\%$ overall
accuracy, only $0.79$ pp below using rendering in both stages
($57.97\%$).
Together with the SFT-only result, this suggests that subsequent RL
enables the model to make more effective use of its SFT initialization,
even without explicit overlays.
One possible explanation is that reference-frame supervision provides
directional knowledge that RL learns to exploit internally for
viewpoint-dependent reasoning.

Explicit rendering is particularly beneficial for hypothetical viewpoints. Using rendering during both RL and evaluation improves Hypo accuracy from $49.40\%$ to $55.66\%$ compared with disabling it in both stages. Even for the model trained without rendering, adding overlays at evaluation increases Hypo accuracy from $49.40\%$ to $55.42\%$, suggesting that explicit reference frames support hypothetical reasoning even without rendering during RL. Since the imagined observer may not be visible in the image, the rendered frame may provide a visual anchor for maintaining consistent reference directions from the hypothetical viewpoint.

\section{Case Study}
\begin{figure*}[t]
    \centering
    \includegraphics[width=\textwidth]{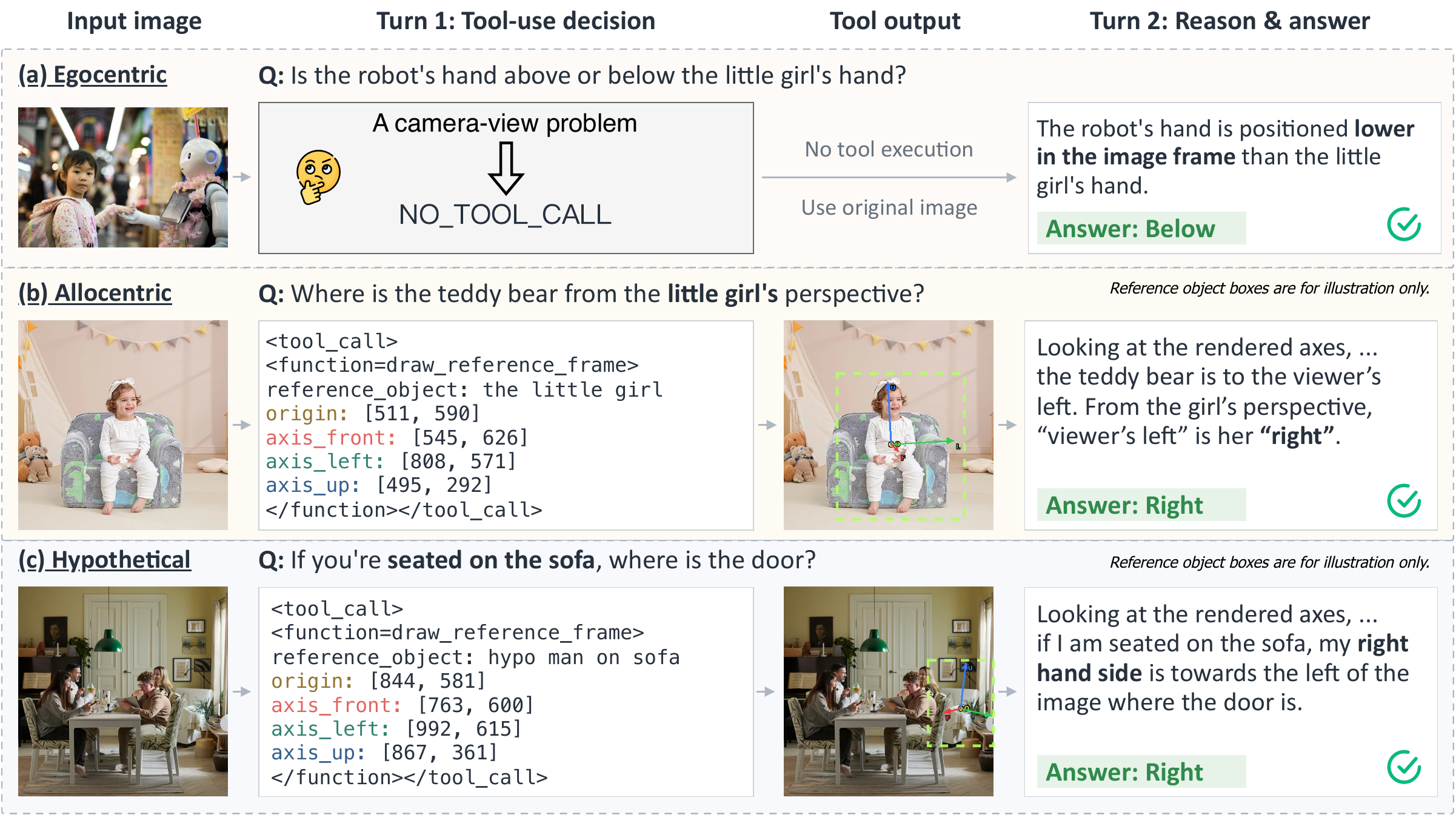}
    \caption{
    Qualitative examples of LeRF-9B on egocentric, allocentric, and hypothetical perspective-taking queries.
    }
    \label{fig:casestudy}
\end{figure*}
Fig.~\ref{fig:casestudy} qualitatively illustrates how LeRF adapts its reasoning strategy to different perspective-taking queries.
For the egocentric example, the queried relation is defined directly in the camera view, and LeRF correctly predicts \texttt{NO\_TOOL\_CALL} and answers from the original image.
In contrast, the allocentric and hypothetical examples require reasoning from the perspective of a scene entity or an imagined observer.
LeRF therefore grounds the corresponding reference entity, constructs its projected reference frame, and revisits the rendered image to infer the requested spatial relation.
These cases demonstrate that LeRF not only learns how to construct useful reference frames, but also when such explicit perspective transformation is necessary.
\section{Discussion and Conclusion}
\paragraph{Discussion.} LeRF provides explicit directional cues without reconstructing a full 3D scene, but its frame predictions can still be affected by occlusion, visual ambiguity, and uncertain entity orientation, which may propagate to downstream reasoning. Moreover, our current formulation focuses on static images, and its effectiveness in dynamic settings where reference frames change over time remains unexplored. Additional failure cases and limitations are discussed in Appendix~\ref{appendix:discussion}.

\paragraph{Conclusion.} We introduced LeRF, a framework that learns to selectively construct and reason with reference frames for perspective taking. Through supervised frame prediction followed by RL, LeRF improves reference frame estimation and perspective taking performance across multiple benchmarks. Our analyses demonstrate the value of rendered directional cues and show that SFT initialization benefits subsequent RL even without explicit frame rendering. These findings highlight learning reference frames as an effective approach to improving viewpoint-dependent reasoning in VLMs. Future work could extend LeRF to dynamic scenes, where reference frames must be tracked and updated as entities move.

\bibliographystyle{alpha}
\bibliography{main}

@inproceedings{zhang2026spinbenchperspectiverotationlens,
  title={Spinbench: Perspective and rotation as a lens on spatial reasoning in vlms},
  author={Zhang, Yuyou and Corcodel, Radu and Hori, Chiori and Cherian, Anoop and Zhao, Ding},
  booktitle={International Conference on Learning Representations},
  volume={2026},
  pages={70072--70141},
  year={2026}
}

@inproceedings{li2025viewspatialbenchevaluatingmultiperspectivespatial,
  title={Viewspatial-bench: Evaluating multi-perspective spatial localization in vision-language models},
  author={Li, Dingming and Li, Hongxing and Wang, Zixuan and Yan, Yuchen and Zhang, Hang and Chen, Siqi and Hou, Guiyang and Jiang, Shengpei and Zhang, Wenqi and Shen, Yongliang and others},
  booktitle={European Conference on Computer Vision},
  pages={95--111},
  year={2026},
  organization={Springer}
}

@article{ma2025spatialreasonerexplicitgeneralizable3d,
  title={Spatialreasoner: Towards explicit and generalizable 3d spatial reasoning},
  author={Ma, Wufei and Chou, Yu-Cheng and Liu, Qihao and Wang, Xingrui and de Melo, Celso and Xie, Jianwen and Yuille, Alan},
  journal={Advances in Neural Information Processing Systems},
  volume={38},
  pages={140751--140774},
  year={2026}
}

@inproceedings{lee2025perspectiveawarereasoningvisionlanguagemodels,
  title={Perspective-aware reasoning in vision-language models via mental imagery simulation},
  author={Lee, Phillip Y and Je, Jihyeon and Park, Chanho and Uy, Mikaela Angelina and Guibas, Leonidas and Sung, Minhyuk},
  booktitle={Proceedings of the IEEE/CVF international conference on computer vision},
  pages={9241--9251},
  year={2025}
}

@article{qwen3.5,
  title={Qwen3. 5-omni technical report},
  author={Team, Qwen},
  journal={arXiv preprint arXiv:2604.15804},
  year={2026}
}

@article{wang2026orientv2unifyingorientation,
  title={Orient anything v2: Unifying orientation and rotation understanding},
  author={Wang, Zehan and Zhang, Ziang and Xu, Jiayang and Wang, Jialei and Pang, Tianyu and Du, Chao and Zhao, Hengshuang and Zhao, Zhou},
  journal={Advances in Neural Information Processing Systems},
  volume={38},
  pages={52622--52645},
  year={2026}
}

@misc{openai2026gpt56,
  title        = {GPT-5.6: Frontier Intelligence that Scales with Ambition},
  author       = {{OpenAI}},
  year         = {2026},
  month        = {july},
  howpublished = {\url{https://deploymentsafety.openai.com/gpt-5-6/}},
  note         = {Official system card}
}

@inproceedings{jia2026omnispatialcomprehensivespatialreasoning,
  title={Omnispatial: Towards comprehensive spatial reasoning benchmark for vision language models},
  author={Jia, Mengdi and Qi, Zekun and Zhang, Shaochen and Zhang, Wenyao and Yu, Xinqiang and He, Jiawei and Wang, He and Yi, Li},
  booktitle={International Conference on Learning Representations},
  volume={2026},
  pages={35634--35670},
  year={2026}
}

@inproceedings{ma20253dsrbenchcomprehensive3dspatial,
  title={3dsrbench: A comprehensive 3d spatial reasoning benchmark},
  author={Ma, Wufei and Chen, Haoyu and Zhang, Guofeng and Chou, Yu-Cheng and Chen, Jieneng and De Melo, Celso and Yuille, Alan},
  booktitle={2025 IEEE/CVF International Conference on Computer Vision (ICCV)},
  pages={6924--6934},
  year={2025},
  organization={IEEE}
}

@article{huang2024rekepspatiotemporalreasoningrelational,
  title={Rekep: Spatio-temporal reasoning of relational keypoint constraints for robotic manipulation},
  author={Huang, Wenlong and Wang, Chen and Li, Yunzhu and Zhang, Ruohan and Fei-Fei, Li},
  journal={arXiv preprint arXiv:2409.01652},
  year={2024}
}

@inproceedings{du2024embspatial,
  title={Embspatial-bench: Benchmarking spatial understanding for embodied tasks with large vision-language models},
  author={Du, Mengfei and Wu, Binhao and Li, Zejun and Huang, Xuan-Jing and Wei, Zhongyu},
  booktitle={Proceedings of the 62nd Annual Meeting of the Association for Computational Linguistics (Volume 2: Short Papers)},
  pages={346--355},
  year={2024}
}

@inproceedings{lee2026multihopspatialmultihopcompositionalspatial,
  title={Multihopspatial: Multi-hop compositional spatial reasoning benchmark for vision-language model},
  author={Lee, Youngwan and Jang, Soojin and Cho, Yoorhim and Lee, Seunghwan and Lee, Yong-Ju and Hwang, Sung Ju},
  booktitle={European Conference on Computer Vision},
  pages={580--598},
  year={2026},
  organization={Springer}
}

@inproceedings{wu2026vtoolr1vlmslearnthink,
  title={Vtool-r1: Vlms learn to think with images via reinforcement learning on multimodal tool use},
  author={Wu, Mingyuan and Yang, Jingcheng and Jiang, Jize and Li, Meitang and Yan, Kaizhuo and Yu, Hanchao and Zhang, Minjia and Zhai, Chengxiang and Nahrstedt, Klara},
  booktitle={International Conference on Learning Representations},
  volume={2026},
  pages={78298--78319},
  year={2026}
}

@article{ma2024imagenet3dgeneralpurposeobjectlevel3d,
  title={Imagenet3d: Towards general-purpose object-level 3d understanding},
  author={Ma, Wufei and Zhang, Guofeng and Liu, Qihao and Zeng, Guanning and Kortylewski, Adam and Liu, Yaoyao and Yuille, Alan},
  journal={Advances in neural information processing systems},
  volume={37},
  pages={96127--96149},
  year={2024}
}

@inproceedings{zhang2024omni6dposebenchmarkmodeluniversal,
  title={Omni6dpose: A benchmark and model for universal 6d object pose estimation and tracking},
  author={Zhang, Jiyao and Huang, Weiyao and Peng, Bo and Wu, Mingdong and Hu, Fei and Chen, Zijian and Zhao, Bo and Dong, Hao},
  booktitle={European Conference on Computer Vision},
  pages={199--216},
  year={2024},
  organization={Springer}
}

@inproceedings{black2023bedlamsyntheticdatasetbodies,
  title={Bedlam: A synthetic dataset of bodies exhibiting detailed lifelike animated motion},
  author={Black, Michael J and Patel, Priyanka and Tesch, Joachim and Yang, Jinlong},
  booktitle={2023 IEEE/CVF conference on computer vision and pattern recognition (CVPR)},
  pages={8726--8737},
  year={2023},
  organization={IEEE}
}

@misc{anthropic2026claudesonnet5,
  title        = {System Card: Claude Sonnet 5},
  author       = {{Anthropic}},
  year         = {2026},
  month        = {june},
  howpublished = {\url{https://www.anthropic.com/claude-sonnet-5-system-card}},
  note         = {Official system card}
}

@article{vteam2026glm45vglm41vthinkingversatilemultimodal,
  title={Glm-4.5 v and glm-4.1 v-thinking: Towards versatile multimodal reasoning with scalable reinforcement learning},
  author={Hong, Wenyi and Yu, Wenmeng and Gu, Xiaotao and Wang, Guo and Gan, Guobing and Tang, Haomiao and Cheng, Jiale and Qi, Ji and Ji, Junhui and Pan, Lihang and others},
  journal={arXiv preprint arXiv:2507.01006},
  year={2025}
}

@article{wang2025internvl35advancingopensourcemultimodal,
  title={Internvl3. 5: Advancing open-source multimodal models in versatility, reasoning, and efficiency},
  author={Wang, Weiyun and Gao, Zhangwei and Gu, Lixin and Pu, Hengjun and Cui, Long and Wei, Xingguang and Liu, Zhaoyang and Jing, Linglin and Ye, Shenglong and Shao, Jie and others},
  journal={arXiv preprint arXiv:2508.18265},
  year={2025}
}

@inproceedings{kaufmann2023emdbelectromagneticdatabaseglobal,
  title={Emdb: The electromagnetic database of global 3d human pose and shape in the wild},
  author={Kaufmann, Manuel and Song, Jie and Guo, Chen and Shen, Kaiyue and Jiang, Tianjian and Tang, Chengcheng and Z{\'a}rate, Juan Jos{\'e} and Hilliges, Otmar},
  booktitle={2023 IEEE/CVF International Conference on Computer Vision (ICCV)},
  pages={14586--14597},
  year={2023},
  organization={IEEE}
}

@inproceedings{krishnan2024omninocsunifiednocsdataset,
  title={Omninocs: A unified nocs dataset and model for 3d lifting of 2d objects},
  author={Krishnan, Akshay and Kundu, Abhijit and Maninis, Kevis-Kokitsi and Hays, James and Brown, Matthew},
  booktitle={European Conference on Computer Vision},
  pages={127--145},
  year={2024},
  organization={Springer}
}

@inproceedings{ahmadyan2020objectronlargescaledataset,
  title={Objectron: A large scale dataset of object-centric videos in the wild with pose annotations},
  author={Ahmadyan, Adel and Zhang, Liangkai and Ablavatski, Artsiom and Wei, Jianing and Grundmann, Matthias},
  booktitle={2021 IEEE/CVF Conference on Computer Vision and Pattern Recognition (CVPR)},
  pages={7818--7827},
  year={2021},
  organization={IEEE}
}

@inproceedings{
zhang2025visionlanguagemodelsrepresentspace,
title={Do Vision-Language Models Represent Space and How? Evaluating Spatial Frame of Reference under Ambiguities},
author={Zheyuan Zhang and Fengyuan Hu and Jayjun Lee and Freda Shi and Parisa Kordjamshidi and Joyce Chai and Ziqiao Ma},
booktitle={The Thirteenth International Conference on Learning Representations},
year={2025},
url={https://openreview.net/forum?id=84pDoCD4lH}
}

@article{ma2026thinkingblueprintsassistingvisionlanguage,
  title={Thinking with blueprints: Assisting vision-language models in spatial reasoning via structured object representation},
  author={Ma, Weijian and Sun, Shizhao and Yu, Tianyu and Wang, Ruiyu and Chua, Tat-Seng and Bian, Jiang},
  journal={arXiv preprint arXiv:2601.01984},
  year={2026}
}

@article{shao2024deepseekmathpushinglimitsmathematical,
  title={Deepseekmath: Pushing the limits of mathematical reasoning in open language models},
  author={Shao, Zhihong and Wang, Peiyi and Zhu, Qihao and Xu, Runxin and Song, Junxiao and Bi, Xiao and Zhang, Haowei and Zhang, Mingchuan and Li, YK and Wu, Yang and others},
  journal={arXiv preprint arXiv:2402.03300},
  year={2024}
}

@article{hu2021loralowrankadaptationlarge,
  title={Lora: Low-rank adaptation of large language models},
  author={Hu, Edward J and Shen, Yelong and Wallis, Phillip and Allen-Zhu, Zeyuan and Li, Yuanzhi and Wang, Shean and Wang, Lu and Chen, Weizhu},
  journal={arXiv preprint arXiv:2106.09685},
  year={2021}
}

@inproceedings{zheng2024llamafactory,
  title={Llamafactory: Unified efficient fine-tuning of 100+ language models},
  author={Zheng, Yaowei and Zhang, Richong and Zhang, Junhao and Ye, Yanhan and Luo, Zheyan},
  booktitle={Proceedings of the 62nd annual meeting of the association for computational linguistics (volume 3: system demonstrations)},
  pages={400--410},
  year={2024}
}

@inproceedings{sheng2024hybridflow,
  title={Hybridflow: A flexible and efficient rlhf framework},
  author={Sheng, Guangming and Zhang, Chi and Ye, Zilingfeng and Wu, Xibin and Zhang, Wang and Zhang, Ru and Peng, Yanghua and Lin, Haibin and Wu, Chuan},
  booktitle={Proceedings of the Twentieth European Conference on Computer Systems},
  pages={1279--1297},
  year={2025}
}

@article{https://doi.org/10.1111/tops.12233,
  title={Comprehending 3D diagrams: Sketching to support spatial reasoning},
  author={Gagnier, Kristin M and Atit, Kinnari and Ormand, Carol J and Shipley, Thomas F},
  journal={Topics in cognitive science},
  volume={9},
  number={4},
  pages={883--901},
  year={2017},
  publisher={Wiley Online Library}
}

@article{yu2017transformations,
  title={Transformations and representations supporting spatial perspective taking},
  author={Yu, Alfred B and Zacks, Jeffrey M},
  journal={Spatial Cognition \& Computation},
  volume={17},
  number={4},
  pages={304--337},
  year={2017},
  publisher={Taylor \& Francis}
}

@article{schultheis2022smart,
  title={A smart model of imaginal perspective taking},
  author={Schultheis, Holger},
  journal={Cognitive Science},
  volume={46},
  number={12},
  pages={e13218},
  year={2022},
  publisher={Wiley Online Library}
}

@article{michelon2006two,
  title={Two kinds of visual perspective taking},
  author={Michelon, Pascale and Zacks, Jeffrey M},
  journal={Perception \& psychophysics},
  volume={68},
  number={2},
  pages={327--337},
  year={2006},
  publisher={Springer}
}

\hfill
\clearpage
\appendix
\newpage
\appendix

\section{Rendering Out-of-Plane Axes}
\label{sec:app-oop-render}

The renderer receives only the four predicted 2D points. Let
$S=\min(W,H)$ and $\hat{\mathbf{a}}_d=\hat{\mathbf{p}}_d-\hat{\mathbf{o}}$
(in pixels). Axis $d$ is treated as out-of-plane if
$\|\hat{\mathbf{a}}_d\|_2<0.02S$. A depth symbol is drawn only if exactly one
axis is out-of-plane and the other two, $\hat{\mathbf{a}}_a$ and
$\hat{\mathbf{a}}_b$, are not nearly collinear, i.e.,
$|\sin\phi|\ge 0.15$ with
$\sin\phi=(\hat{\mathbf{a}}_a\times\hat{\mathbf{a}}_b)/
(\|\hat{\mathbf{a}}_a\|_2\|\hat{\mathbf{a}}_b\|_2)$,
where $(a,b)$ is ordered such that $\mathbf{e}_d=\mathbf{e}_a\times\mathbf{e}_b$
in our right-handed frame. Since image coordinates ($u$ rightward, $v$
downward) and the camera's forward axis form a right-handed system, the sign
of $\sin\phi$ gives the depth direction: we draw $\odot$ (toward the camera)
if $\sin\phi<0$ and $\otimes$ (away from the camera) otherwise, replacing the
arrow. In all other cases, the short arrows are drawn as-is.

In our labels, this corresponds to axes within approximately $3.8^\circ$ of
the viewing direction.

\section{Details of SFT Data Construction}
\label{appendix:sft-data}
\subsection{Reference Frame Label Construction}
\label{appendix:frame-label-construction}

We construct projected reference frames using the first-order projection
described in the main text. A shared scale is applied to all three axes
to preserve their relative projected lengths, chosen so that an axis
parallel to the image plane spans $0.3\min(W,H)$ pixels, where $W$ and $H$
denote the image width and height. If any endpoint falls outside the
image, we uniformly shorten all axes until they fit within the image
bounds. Samples whose frame origin lies outside the image are discarded.
Finally, horizontal and vertical coordinates are normalized by the image
width and height, respectively, and quantized to integers in $[0,1000]$.
As shown in Appendix~\ref{sec:length-of-axis-exp}, the exact axis length
has little effect once the axes are sufficiently long.

\subsection{System Prompt of LeRF Training and Inference}
\label{appendix:system_prompt}
\begin{tcolorbox}[
    colback=white,
    colframe=gray,
    title=System Prompt of LeRF Training and Inference,
    fonttitle=\large,
    arc=4mm,
    breakable
]
You are a vision-language assistant specialized in spatial and perspective-taking reasoning. You have access to the visual tool `draw\_reference\_frame`. \\

Your goal is to solve spatial questions accurately and to use `draw\_reference\_frame` only when perspective transformation is necessary.\\

For each question, select the appropriate first-turn action based on whether it can be answered directly from the current camera/image perspective, or whether it requires reasoning from the perspective or orientation of a particular object, person, or hypothetical observer.\\

- If the question can be answered correctly from the camera/image perspective without changing the reference frame, do not call `draw\_reference\_frame`. On the first assistant turn, output exactly the literal string `NO\_TOOL\_CALL` and nothing else. `NO\_TOOL\_CALL` is an intermediate control signal, not the final answer. After receiving a continuation instruction, reason from the original image and produce the final answer.
- If the question requires a perspective transformation or object-centered reasoning, you must call `draw\_reference\_frame` before answering, explicitly construct the corresponding object-centered reference frame, and use the returned annotated image for subsequent reasoning.\\

The first assistant response MUST contain exactly one of the following:\\

1. the literal string `NO\_TOOL\_CALL`, if perspective transformation is not required;
2. a single call to `draw\_reference\_frame`, if perspective transformation is required.\\

Do not output explanations, reasoning, or the final answer on the first assistant turn. Perform the first-turn routing action only once for each original question. After receiving either a tool result or a continuation instruction, do not repeat the routing action and proceed with the corresponding subsequent reasoning steps.\\

The mere presence of an object, person, or spatial-direction word in the question does not by itself require a call to `draw\_reference\_frame`. Call `draw\_reference\_frame` only when the correct interpretation of the spatial relation depends on the perspective or orientation of a particular reference entity.\\

\#\# Reference-frame convention\\

A reference frame is a 3D object-centered coordinate frame projected onto the 2D image plane.\\

For a reference entity:\\

* `+X` is its intrinsic **front** direction.
* `+Y` is its intrinsic **left** direction.
* `+Z` is its intrinsic **up** direction.\\

These directions are defined relative to the reference entity itself, not relative to the camera or the image.\\

The frame is right-handed:

`+X × +Y = +Z`

or equivalently:

`front × left = up`\\

The corresponding negative directions are:

* **back** = `-X`;
* **right** = `-Y`;
* **down** = `-Z`.

The projected frame is represented by four absolute image points:\\

* `O`: the projected frame origin;\\
* `axis\_front`: the projected endpoint of the `+X` / front axis;\\
* `axis\_left`: the projected endpoint of the `+Y` / left axis;\\
* `axis\_up`: the projected endpoint of the `+Z` / up axis.\\

Use `O` consistently when referring conceptually to the frame origin. Use `origin` only as the corresponding function parameter name.\\

All coordinates are integers in `[0, 1000]`, with the image origin at the top-left. `x` increases to the right and `y` increases downward.\\

The axis endpoints are absolute image points, not direction vectors. The displacement from `O` to an endpoint is the visible 2D projection of the corresponding positive 3D direction.\\

\#\# Rendered-frame semantics\\

`draw\_reference\_frame` renders the predicted frame onto the image and returns the annotated image.\\

The rendered directions use the following colors:

* red = front / `+X`;
* green = left / `+Y`;
* blue = up / `+Z`.\\

The rendered image may also contain the following labels:

* `F` = front;
* `L` = left;
* `U` = up;
* `O` = frame origin.\\

When an axis is too strongly foreshortened to be displayed reliably as an arrow, the tool replaces that axis's arrow with a depth marker drawn at `O`:\\

* `$\odot$` means that the positive axis points toward the camera and out of the image plane;
* `$\otimes$` means that the positive axis points away from the camera and into the image plane.\\

A depth marker uses the color and label of its corresponding axis, and is drawn instead of that axis's arrow, so a marked axis has a marker and a label but no arrow. The other two axes are still drawn as arrows.\\

At most one depth marker appears in an image. The tool draws one only when exactly one axis is strongly foreshortened and the other two are clearly non-collinear, because the marker's direction is recovered from those two axes. If two or three axes collapse, or the remaining two are nearly collinear, no marker is drawn; in that case do not assume a depth direction, and treat the frame as unreliable rather than inventing one.\\

Depth markers are generated automatically by the tool and are not additional function-call parameters.\\

Treat all colors, labels, arrows, origin indicators, and depth markers as components of the rendered reference frame, not as scene objects or additional spatial entities.\\

\#\# Required procedure\\

\#\#\# 1. Determine whether perspective transformation is necessary and identify the reference entity if needed\\

First determine whether the question can be answered directly from the camera/image perspective or requires adopting the perspective or orientation of a particular entity.\\

If the question can be answered directly from the camera/image perspective, do not identify or invent a reference entity and do not call `draw\_reference\_frame`. Output exactly `NO\_TOOL\_CALL` and stop the current turn. After receiving a continuation instruction, skip Steps 2–5 and proceed to Step 6 using the original image.\\

If perspective transformation is required, determine whose or which entity's perspective or orientation is required by the question.\\

For an explicitly mentioned observer or object, use that actual entity as the reference entity. For example:

* `the long-haired player`;
* `the cyclist in white`;
* `the woman holding the umbrella`;
* `the red office chair by the window`.\\

Do not confuse the reference entity with another object that is merely mentioned in the question.\\

For a hypothetical-viewpoint expression such as:\\

* “if you are standing in front of the refrigerator”;
* “suppose someone is standing in front of the table”;
* “from the perspective of someone standing in front of the desk”;\\

use the hypothetical observer as the reference entity and infer its position and orientation from the relation specified in the question. Describe it unambiguously, for example:\\

* `the hypothetical observer standing in front of the refrigerator`;
* `the hypothetical observer standing in front of the table`.\\

If the question does not require adopting the perspective or orientation of any entity, follow the `NO\_TOOL\_CALL` procedure above rather than inventing a reference frame.\\

If a later reasoning step requires a different reference entity, construct a new frame for that entity before using its perspective.\\

\#\#\# 2. Estimate the projected reference frame, if you decide to call the tool `draw\_reference\_frame`\\

Locate the reference entity or its hypothetical position in the image.\\

Infer its intrinsic front, left, and up directions, and predict:\\

* the frame origin `O`;
* the projected front-axis endpoint;
* the projected left-axis endpoint;
* the projected up-axis endpoint.\\

Use the reference-frame and image-coordinate conventions defined above.\\

Only place an axis endpoint at or near `O` when perspective projection genuinely makes that axis strongly foreshortened. Never collapse an axis merely because its direction is uncertain.\\

\#\#\# 3. Call `draw\_reference\_frame`, if it is necessary for the problem\\

Call `draw\_reference\_frame` with the estimated reference entity and frame coordinates.\\

When a reference frame is required:\\

* do not answer the spatial question before obtaining the tool result;
* the first assistant response MUST contain only the tool call;
* do not output explanations, reasoning, or any other text before or after that first tool call.\\

\#\#\# 4. Inspect the returned image, if the `draw\_reference\_frame` is called\\

Inspect the annotated image and identify the rendered:\\

* origin `O`;
* front direction `F`;
* left direction `L`;
* up direction `U`;
* any `$\odot$` or `$\otimes$` depth markers.\\

Use colors, labels, and marker shapes together to distinguish the rendered directions.\\

\#\#\# 5. Reason in the rendered reference frame, if the `draw\_reference\_frame` is called\\

Treat the rendered frame in the returned image as the authoritative representation of the reference entity's orientation.\\

The rendered axes are perspective projections of 3D directions. They need not be perpendicular or equally long in the image, and two axes may overlap or become nearly collinear. Preserve their identities using their colors, labels, and depth markers; do not rearrange, swap, or orthogonalize them. Do not interpret projected axis length as physical distance or confidence.\\

A strongly foreshortened axis remains a valid 3D direction even when it is represented by a depth marker instead of an arrow. Use the marker's defined depth direction rather than inventing an arbitrary image-plane direction.\\

Use the rendered frame to define orientation, while using the annotated scene to identify other entities and infer their relative positions and depth. The frame provides a local 3D orientation at `O`, not a global 2D Cartesian grid across the image. Therefore, do not determine spatial relations solely from image-left/image-right, raw image coordinates, projected axis length, or a simple 2D dot product.\\

Do not independently re-derive or override the reference entity's orientation from object appearance, camera-centric relations, linguistic priors, or memorized object orientations. If any of these conflict with the rendered frame, follow the rendered frame.\\

\#\#\# 6. Reason with the appropriate visual input and produce the final answer\\

If the first-turn response was `NO\_TOOL\_CALL` and a continuation instruction has been received, reason directly from the original image using the camera/image perspective.\\

If `draw\_reference\_frame` was called, reason from the returned annotated image using the rendered reference frame as specified above.
\\
After completing the appropriate reasoning, answer the user's question directly and concisely.\\

The final answer MUST be enclosed in `\texttt{\textbackslash boxed\{\}}`. Put only the final answer itself inside `\texttt{\textbackslash boxed\{\}}`.\\

Examples:\\

* `\texttt{\textbackslash boxed\{left\}}` \\
* `\texttt{\textbackslash boxed\{B\}}`\\
* `\texttt{\textbackslash boxed\{the white mug\}}`\\

This requirement applies only to the final answer after the first-turn action has been handled. Never output `\texttt{\textbackslash boxed\{\}}` as part of a `draw\_reference\_frame` tool-call response or a `NO\_TOOL\_CALL` response.\\

Do not expose intermediate coordinate estimates, function arguments, or other hidden intermediate information in the final answer unless the user explicitly requests them.\\

\#\# Important constraints\\

- The first assistant response MUST contain exactly one action and no other text: output `NO\_TOOL\_CALL` if perspective transformation is unnecessary, or output a single `draw\_reference\_frame` tool call if perspective transformation is required. Never output the final answer on the first assistant turn.\\

- Never equate image-left/image-right with the reference entity's left/right, or the camera's viewing direction with the reference entity's front/back.\\

- Treat the rendered O, F, L, U, $\odot$, and $\otimes$ indicators as authoritative components of the reference frame, not as scene objects. Do not independently re-estimate or override the rendered orientation.\\

- Interpret the rendered frame as a perspective projection of a 3D object-centered frame, not as a flat 2D Cartesian grid.\\

- Never collapse an axis merely because its direction is uncertain. An endpoint may appear at or near O only because of genuine perspective foreshortening.\\

- `NO\_TOOL\_CALL` is an intermediate control signal indicating that the original camera/image perspective should be used. It is not a final answer and must never be enclosed in `\texttt{\textbackslash boxed\{\}}`.\\

- The final response MUST contain only the requested answer inside `\texttt{\textbackslash boxed\{\}}`. Do not expose intermediate coordinates or tool arguments.\\
\end{tcolorbox}

\subsection{Object Categories adopted in SFT training}
\label{sec:filtered-obj-category-for-sft}
\begin{tcolorbox}[
    colback=white,
    colframe=gray,
    title=Adopted Object Categories in ImageNet3D,
    fonttitle=\large,
    arc=4mm,
    breakable
]
car, aeroplane, boat, bed, tvmonitor, stove, motorbike, sofa, bicycle, bus, chair, bench, laptop, printer, refrigerator, microwave, toilet, forklift, ambulance, fire truck, school bus, tractor, train.
\end{tcolorbox}

\begin{tcolorbox}[
    colback=white,
    colframe=gray,
    title=Adopted Object Categories in Omni6DPose-SOPE,
    fonttitle=\large,
    arc=4mm,
    breakable
]
mug, teapot, book, keyboard, calculator, mouse, remote control, shoe, knife, toy bus, toy truck, toy train, shrimp, toy boat, brush, toy plane, hair dryer, helmet, shampoo, camera, toy motorcycle, toy animals, laptop, spoon, dinosaur, fork, garage kit, laundry detergent, fire extinguisher, kettle, projector, backpack, doll, teddy bear, ricecooker, tooth brush, glasses, pan, razor, microwave oven, spanner.
\end{tcolorbox}

\subsection{Effect of Coordinate Aaxis Length} 
\label{sec:length-of-axis-exp}
\begin{figure*}[t]
    \centering
    \includegraphics[width=0.6\textwidth]{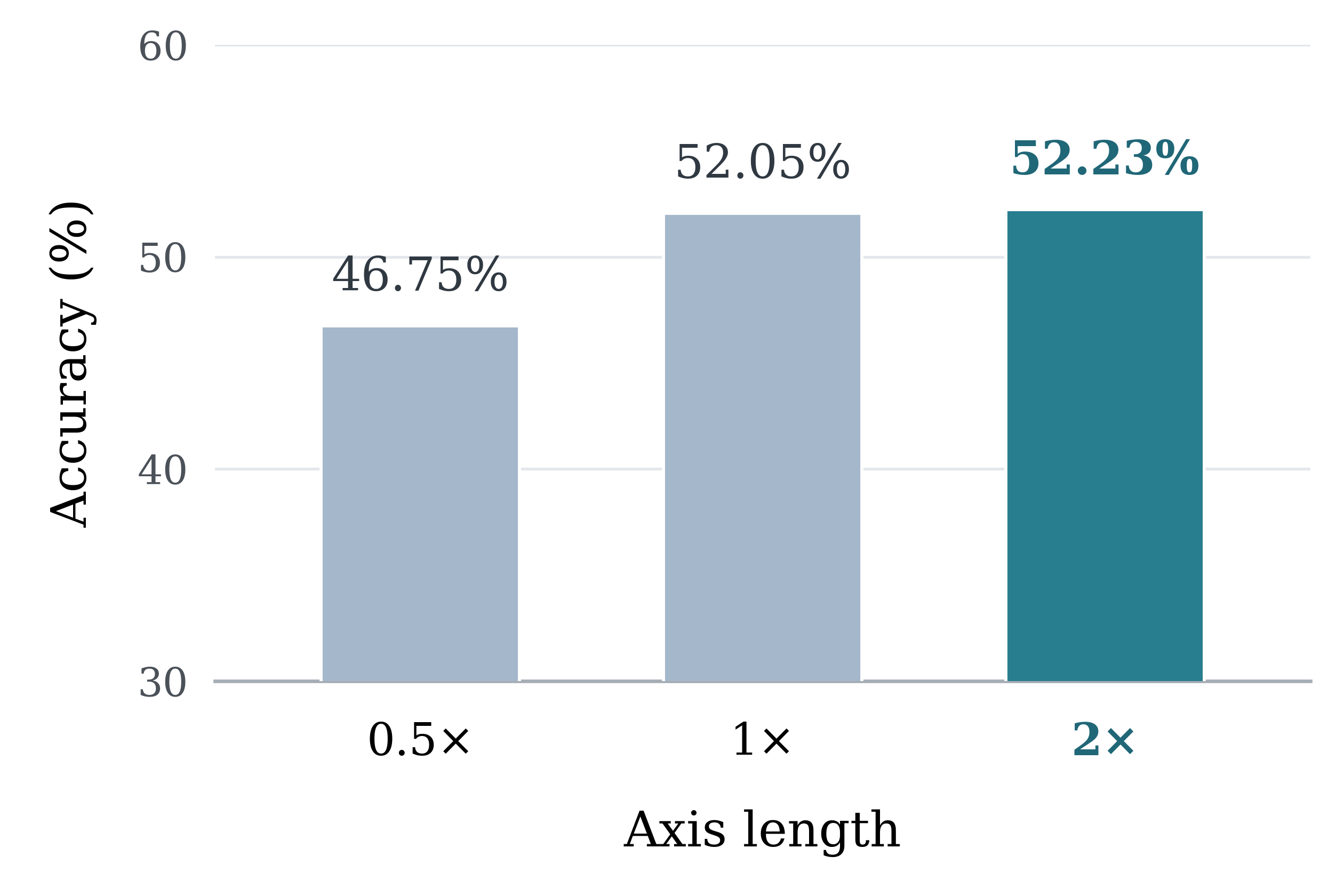}
    \caption{
Effect of coordinate-axis length on accuracy.
    }
    \label{fig:axis_length}
\end{figure*}

We pick 415 cases in OmniSpatial-Bench \citep{jia2026omnispatialcomprehensivespatialreasoning} which requires perspective changing and has real reference entity and use OrientAnything-V2 \citep{wang2026orientv2unifyingorientation} to overlay reference frame to the image. We then manually resize the reference frame to $0.5\times$ and $2\times$ and run evaluations on Qwen3.5-9B (thinking) with these three axes length settings. The result is shown in Table \ref{fig:axis_length}.

\clearpage
\subsection{Question Templates for Tool-Calling Data}
\label{sec:qtemplate-tool-calling}
\begin{tcolorbox}[
    colback=white,
    colframe=gray,
    title=Question Templates for Tool-Calling Data,
    fonttitle=\large,
    arc=4mm,
    breakable,
    left=4mm,
    right=4mm,
    top=2mm,
    bottom=2mm
]
\small
\begin{itemize}
    \setlength{\itemsep}{0.25em}
    \setlength{\parskip}{0pt}
    \setlength{\parsep}{0pt}

    \item From the perspective of \{obj\}, which object in the image is closest to its left?

    \item From the perspective of \{obj\}, how many objects are on its right side?

    \item From \{obj\}'s point of view, is the object nearest to the camera on its left or right?

    \item If I stand at \{obj\}'s position and face the same direction it faces, which object is directly in front of me?

    \item If you were standing where \{obj\} is and facing the same direction, would the camera be in front of you, behind you, on your left, or on your right?

    \item Which object in the image is \{obj\} facing toward?

    \item Which side of \{obj\}---its front, back, left, or right---is facing the camera?

    \item From the perspective of \{obj\}, which is the tallest object behind it?

    \item From the perspective of \{obj\}, is the largest object in the scene in front of it or behind it?

    \item If I stand at \{obj\}'s position, face the same direction it faces, and then turn $90^\circ$ to my left, which object will be directly in front of me?
\end{itemize}
\end{tcolorbox}

\subsection{Question Templates for No Tool-Calling Data}
\label{sec:qtemplate-no-tool-calling}
\begin{tcolorbox}[
    colback=white,
    colframe=gray,
    title=Question Templates for No Tool-Calling Data,
    fonttitle=\large,
    arc=4mm,
    breakable,
    left=4mm,
    right=4mm,
    top=2mm,
    bottom=2mm
]
\small
\begin{itemize}
    \setlength{\itemsep}{0.25em}
    \setlength{\parskip}{0pt}
    \setlength{\parsep}{0pt}

    \item Which object is the closest to the \{obj\}?

    \item From the camera's point of view, how many objects appear to the right of the \{obj\} in the image?

    \item From the photographer's point of view, is the \{obj\} on the left or on the right side of the image?

    \item In this image, which object appears directly above the \{obj\}?

    \item In the picture, is the \{obj\} in the top-left, top-right, bottom-left, or bottom-right region of the image?

    \item Which object is the \{obj\} next to?

    \item Which side of the image---the top, the bottom, the left, or the right---is the \{obj\} closest to?

    \item Is the \{obj\} the highest and biggest object in this image?

    \item Judging from the camera's viewpoint, is the largest object in the scene closer to the camera than the \{obj\}?

    \item How many \{obj\} are there in the image?
\end{itemize}
\end{tcolorbox}

\section{Details of Experiments}
\subsection{LeRF Training Details}
\label{sec:training-details}

LeRF-4B and LeRF-9B use the same training configuration, initialized from Qwen3.5-4B and Qwen3.5-9B \citep{qwen3.5}, respectively. Both training stages use four NVIDIA RTX PRO 6000 Blackwell GPUs (96GB each), with the vision encoder and multimodal projector frozen. Details are listed in Table \ref{tab:train_hparams}.

\begin{table}[h]
\centering
\caption{Detailed training hyperparameters.}
\label{tab:train_hparams}
\resizebox{0.9\linewidth}{!}{
\begin{tabular}{@{}lll@{}}
\toprule
Config                  & SFT                       & RL (GRPO) \\
\midrule
Trainable parameters    & LoRA ($r{=}8$, $\alpha{=}16$, all linear; vision tower \& projector frozen) & Full LLM \\
Max image pixels        & $1024 \times 1024$        & -- \\
LR scheduler            & Cosine with warm-up       & Constant \\
Warm-up                 & 10\% of steps             & None \\
Optimizer               & AdamW ($\beta_1{=}0.9$, $\beta_2{=}0.999$, $\epsilon{=}10^{-8}$) & AdamW ($\beta_1{=}0.9$, $\beta_2{=}0.999$) \\
Global batch size       & 64                        & 64 prompts $\times$ 8 rollouts \\
PPO mini-batch size     & --                        & 32 prompts \\
Learning rate           & $1 \times 10^{-4}$        & $1 \times 10^{-6}$ \\
Weight decay            & 0                         & $1 \times 10^{-2}$ \\
Gradient clipping       & 1.0                       & 1.0 \\
Training precision      & \textit{bfloat16}         & \textit{bfloat16} \\
Parallelism             & DDP (4 GPUs)              & FSDP2 w/ param \& optimizer offload (4 GPUs) \\
Epochs                  & 2                         & 1 \\
Max sequence length     & 6400                      & 4608 (prompt) + 12288 (response) \\
Thinking budget         & --                        & 10240 tokens \\
\midrule
Number of rollouts $K$  & --                        & 8 \\
Rollout temperature     & --                        & 1.0 (train) / 0.6 (val) \\
KL penalty $\beta$      & --                        & 0.01 (low-variance KL, as loss) \\
Advantage normalization & --                        & Group mean \& std \\
PPO clip ratio          & --                        & 0.2 (dual-clip $c{=}3.0$) \\
Loss aggregation        & --                        & Token mean \\
Reward                  & --                        & Binary accuracy (requires exactly one \texttt{\textbackslash boxed\{\}}) \\
\bottomrule
\end{tabular}
}
\end{table}


\subsection{Latency Analysis}
\label{appendix:latency}
\begin{table}[t]
\centering
\caption{
Inference efficiency comparison.
All models are measured under single-request inference
on an RTX PRO 6000 GPU.
}
\label{tab:latency}
\setlength{\tabcolsep}{6pt}
\begin{tabular}{lccc}
\toprule
Model
& Output Tokens $\downarrow$
& Latency (s) $\downarrow$
& Decoding Speed \\
\midrule
SpatialReasoner-7B
& 317
& 2.2
& 68 tok/s \\

LeRF-9B
& 3757
& 51.5
& 77 tok/s \\

Qwen3.5-9B Thinking
& 6603
& 77.0
& 81 tok/s \\

APC-Qwen3.5-9B Thinking
& 26140
& 367.5
& 71 tok/s \\
\bottomrule
\end{tabular}
\end{table}
We additionally compare inference latency and output length across different methods. As shown in Table~\ref{tab:latency}, LeRF-9B produces substantially shorter outputs than reasoning-intensive baselines, averaging 3.8K tokens per question compared with 6.6K for Qwen3.5-9B with thinking and 26.1K for APC-Qwen3.5-9B. Correspondingly, its average latency is 51.5\,s, compared with 77.0\,s and 367.5\,s, respectively. Since all models exhibit similar decoding throughput (approximately 70--80 tokens/s), the latency difference is largely explained by output length. In particular, LeRF reduces both output length and latency by roughly one third relative to Qwen3.5-9B with thinking, while requiring only about one seventh of the computation of APC. All latency experiments are measured on the same RTX PRO 6000 GPU under single-request inference. 

\subsection{Additional Qualitative Comparison with Baselines}
\label{appendix:more_case_study}
\begin{figure*}[t]
    \centering
    \includegraphics[width=\textwidth]{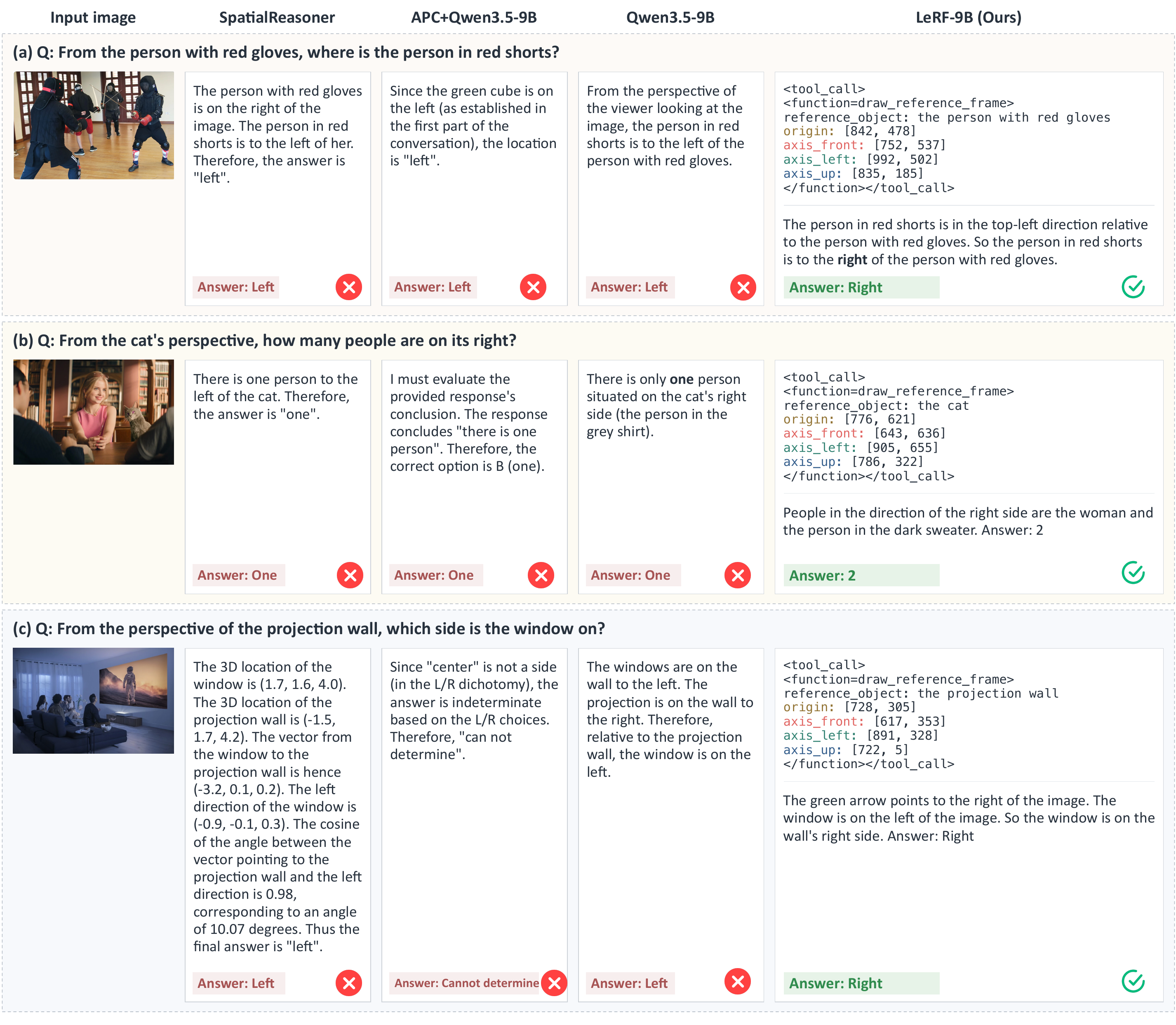}
    \caption{
Additional qualitative comparison with SpatialReasoner, APC+Qwen3.5-9B, and Qwen3.5-9B.
Baseline models frequently answer according to camera-centric spatial relations or fail to resolve the requested viewpoint, whereas LeRF explicitly constructs an entity-centered reference frame and correctly reasons from the specified perspective.
    }
    \label{fig:more_case_study}
\end{figure*}

Figure~\ref{fig:more_case_study} provides additional qualitative comparisons between LeRF and several representative baselines on perspective-taking questions. Across these examples, the baseline models often reason directly from the camera or image coordinate system, even when the query explicitly specifies the viewpoint of another entity.

In Figure~\ref{fig:more_case_study}(a), all three baselines judge the person in red shorts to be on the left because this relation holds from the camera viewpoint. LeRF instead constructs a reference frame for the person with red gloves and correctly resolves the direction as \textit{right} from that person's perspective. A similar pattern appears in Figure~\ref{fig:more_case_study}(b): the baselines count only one person on the cat's right, whereas the reference frame predicted by LeRF makes the cat-centered right direction explicit and leads to the correct count of two. In Figure~\ref{fig:more_case_study}(c), the baselines either follow the apparent left--right relation in the image or fail to determine the answer. By grounding the projection wall's intrinsic orientation, LeRF correctly identifies the window as lying on its \textit{right} side.

These examples further illustrate that perspective-taking errors are not necessarily caused by an inability to recognize the relevant entities or their image-space locations. Rather, a key difficulty is transforming these observations into the coordinate system specified by the query. Explicit reference-frame construction provides LeRF with a concrete intermediate representation for performing this viewpoint transformation.

\section{Further Discussion}
\label{appendix:discussion}
\subsection{Extended Discussion of Limitations}
LeRF intentionally uses a lightweight projected reference frame rather than reconstructing a full 3D scene. While this representation is sufficient for many viewpoint-dependent directional relations, it does not explicitly encode metric depth, scale, or complete scene geometry. Tasks that require precise 3D distances, heavily occluded relations, or more complex geometric transformations may therefore benefit from richer spatial representations.

Reliable frame construction also assumes that the reference entity has a visually identifiable intrinsic orientation. Heavy occlusion, ambiguous appearance, or entities without a well-defined front or left direction can make the predicted frame unreliable. This limitation is also reflected in our training data construction, where object categories without meaningful intrinsic directions are excluded. Finally, our current formulation operates on static images. Extending LeRF to videos would require reference frames to be tracked and updated over time as entities and viewpoints change, while avoiding the accumulation of frame-estimation errors.

\subsection{Failure Cases Analysis}
\begin{figure*}[t]
    \centering
    \includegraphics[width=\textwidth]{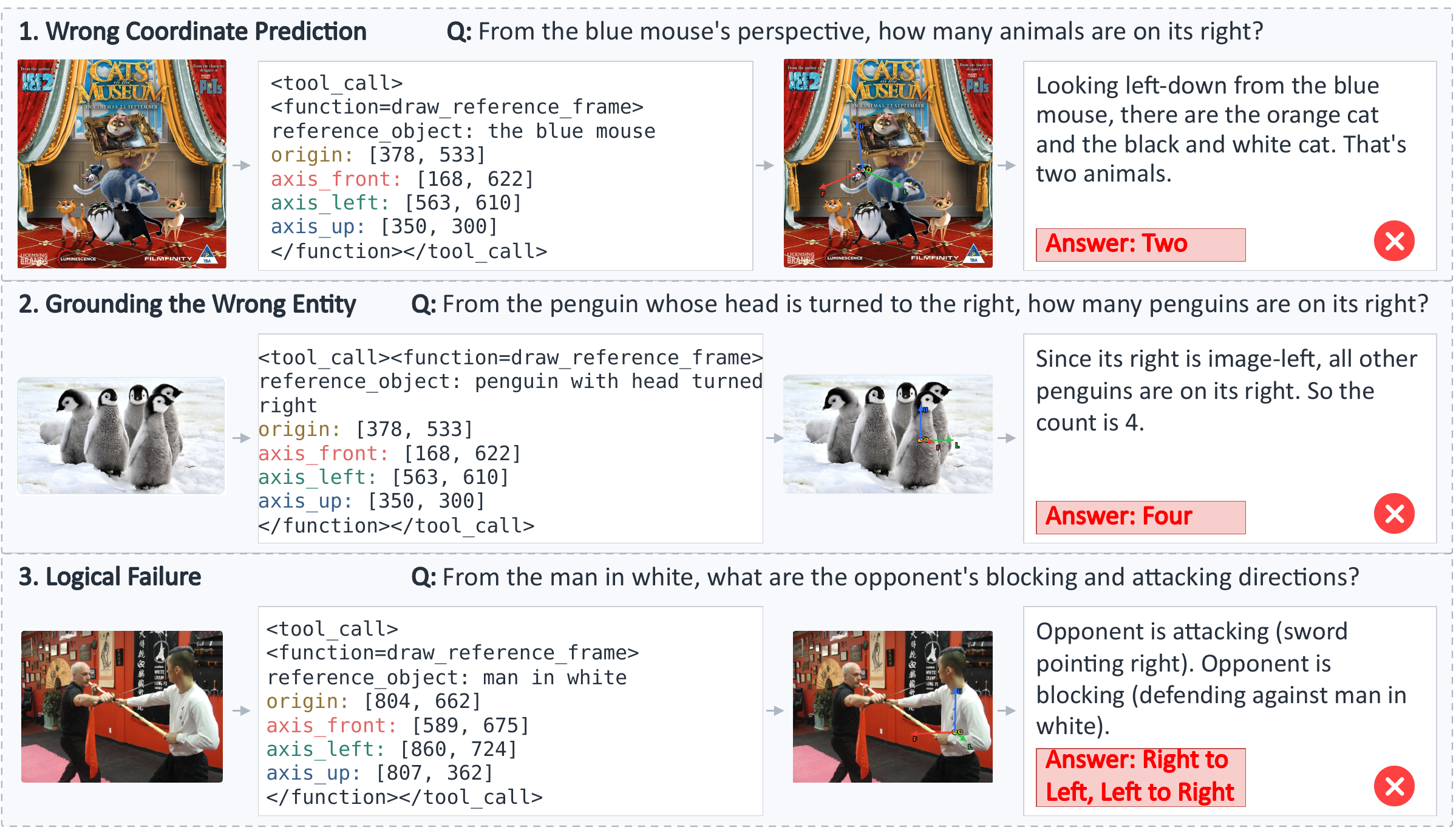}
    \caption{
    Representative failure cases of LeRF.
Errors can arise from inaccurate reference-frame prediction, incorrect grounding of the reference entity, or downstream reasoning failures even when the reference frame is reasonably constructed.
    }
    \label{fig:failure_cases}
\end{figure*}

Figure~\ref{fig:failure_cases} presents representative failure cases of LeRF.
We observe three major sources of error.
First, the model may identify the correct reference entity but predict an inaccurate reference frame, leading to an incorrect transformation of viewpoint-dependent directions.
Second, it may ground the wrong reference entity, causing the subsequently constructed frame to be attached to an unintended object.
Third, even when the reference entity and reference frame are reasonably identified, the model can still fail during downstream reasoning, for example by incorrectly interpreting the queried spatial relation.

In addition to these reference-frame-related failures, LeRF inherits general reasoning limitations of the underlying VLM.
For example, we observe occasional errors in counting, object identification, and multi-step spatial reasoning.
These cases suggest that explicit reference frames can alleviate errors caused by implicit viewpoint transformation, but do not eliminate failures originating from the underlying visual perception and reasoning capabilities of the model.


\end{document}